\pdfoutput=1
\documentclass{article}

\usepackage[main,final]{neurips_2026}

\usepackage[utf8]{inputenc}
\usepackage[T1]{fontenc}
\usepackage{hyperref}
\usepackage{url}
\usepackage{booktabs}
\usepackage{amsfonts}
\usepackage{amsmath}
\usepackage{nicefrac}
\usepackage{microtype}
\usepackage[table]{xcolor}
\usepackage{graphicx}
\usepackage{subcaption}
\usepackage{algorithm}
\usepackage{algorithmic}
\usepackage{float}
\usepackage{multirow}
\usepackage{arydshln}
\usepackage{enumitem}
\usepackage{wrapfig}
\usepackage{pifont}
\usepackage[most]{tcolorbox}
\usepackage{fancyvrb}
\usepackage{fvextra}
\newcommand{\cmark}{\ding{51}}
\newcommand{\xmark}{\ding{55}}
\usepackage{makecell}
\usepackage{hyperref}
\hypersetup{colorlinks,citecolor={cadmiumgreen}}
\definecolor{cadmiumgreen}{rgb}{0.0,0.42,0.24}
\newtcolorbox{promptbox}[1][]{%
  enhanced, breakable,
  colback=gray!3,
  colframe=gray!80,
  colbacktitle=gray!80,
  coltitle=white,
  fonttitle=\bfseries,
  title={#1},
  top=1mm, bottom=1mm, left=1.5mm, right=1.5mm
  }

\definecolor{line-blue}{RGB}{230, 240, 255}

\newcommand{\ours}{VLM-VGM Collaborative Video Reasoning}
\newcommand{\oursshort}{\textbf{CollabVR}}
\newcommand{\moduleone}{VLM-Driven Progressive Planning}
\newcommand{\moduletwo}{VLM-VGM Collaborative Reasoning}

\newcommand{\vbvrwan}{VBVR-Wan2.2}
\newcommand{\veo}{Veo~3.1}
\newcommand{\sora}{Sora~2}
\newcommand{\gemini}{Gemini~2.5~Pro}

\newcommand{\genvire}{Gen-ViRe}
\newcommand{\vbvr}{VBVR-Bench}

\setcitestyle{numbers,square,comma,sort&compress}

\makeatletter
\renewcommand{\@bottomtitlebar}{%
  \vskip 0.29in
  \vskip -\parskip
  \hrule height 1\p@
  \vskip 0.09in
  \vspace{-0.8em}%
}
\makeatother

\title{CollabVR: Collaborative Video Reasoning with Vision-Language and Video Generation Models}

\author{%
  Joowon Kim\textsuperscript{1}\thanks{Equal contribution.\quad $^{\dagger}$Corresponding author.}\hspace{1.2em}%
  Seungho Shin\textsuperscript{1}\footnotemark[1]\hspace{1.2em}%
  Eunho Yang\textsuperscript{1,2}\hspace{1.2em}%
  Joonhyung Park\textsuperscript{1,3$\dagger$}\\[2pt]
  \textsuperscript{1}KAIST \quad \textsuperscript{2}AITRICS \quad \textsuperscript{3}Kyung Hee University\\[2pt]
  \texttt{\small \{kjwispro, ssh9918, eunhoy, deepjoon\}@kaist.ac.kr}%
}

\begin{document}

\maketitle

\begin{figure}[!h]
    \centering
    \vspace{-2em}
    \includegraphics[width=\linewidth]{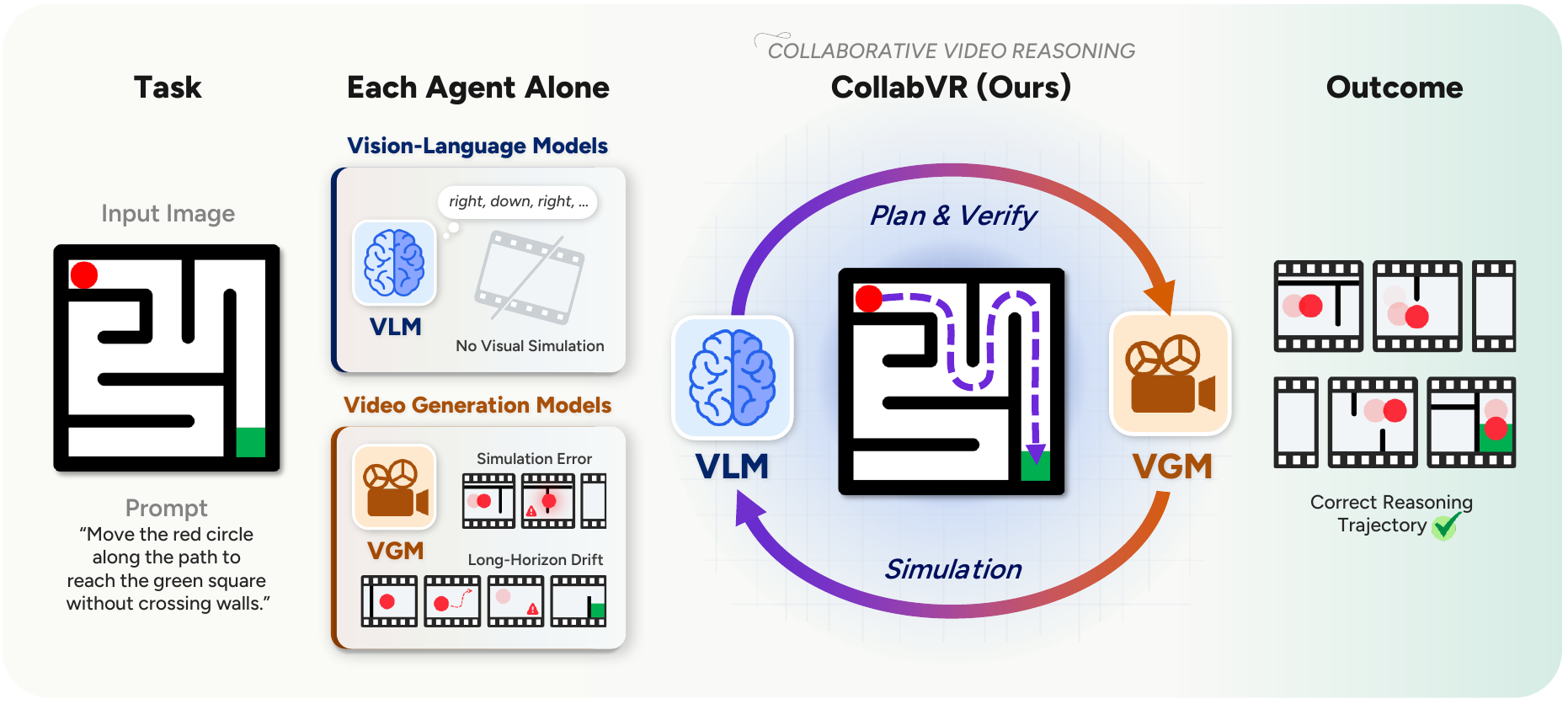}
    \caption{
        \textbf{VLM as planner, VGM as simulator.}
        A VLM is strong at reasoning but weak at visual simulation, while a VGM simulates short clips but lacks reasoning, causing long-horizon drift and mid-clip simulation errors.
        \oursshort{} couples them in a closed loop where the VLM plans progressively and diagnoses each generated clip, turning failures into correctable signals.
    }
    \label{fig:concept}
\end{figure}

\begin{abstract}
Recent \textit{Thinking with Video} approaches use Video Generation Models (VGMs) for visual reasoning by producing temporally coherent Chain-of-Frames as reasoning artifacts.
Even strong VGMs, however, exhibit two recurring failure modes on goal-directed tasks: long-horizon drift on multi-step tasks and mid-clip simulation errors that compound.
Both stem from the absence of explicit reasoning built upon the VGM's short-horizon visual prior, a role naturally filled by Vision-Language Models (VLMs), but where to place the VLM is non-trivial: upfront plans commit before any frame is generated and post-hoc critiques over whole videos intervene too late.
We propose \ours{} (\oursshort), a closed-loop framework that couples the VLM with the VGM at step-level granularity: the VLM plans the immediate next action, inspects the clip the VGM generates, and folds the verifier's diagnosis directly into the next action prompt to repair detected failures.
On \genvire{} and \vbvr{}, \oursshort{} improves both open-source and closed-source VGMs over single-inference, Pass@$k$, and prior test-time scaling baselines at matched compute, with the largest gains on the hardest tasks.
It also yields further improvements on top of a reasoning-fine-tuned VGM, indicating that step-level VLM supervision is orthogonal to and stackable with reasoning-oriented fine-tuning.
We provide video samples and additional qualitative results at our project page: \url{https://joow0n-kim.github.io/collabvr-project-page/}.
\end{abstract}

\section{Introduction}
\label{sec:introduction}

\setlength{\intextsep}{4pt}
\setlength{\columnsep}{10pt}
\begin{wrapfigure}{r}{0.4\linewidth}
    \centering
    \vspace{-3em}
    \includegraphics[width=\linewidth]{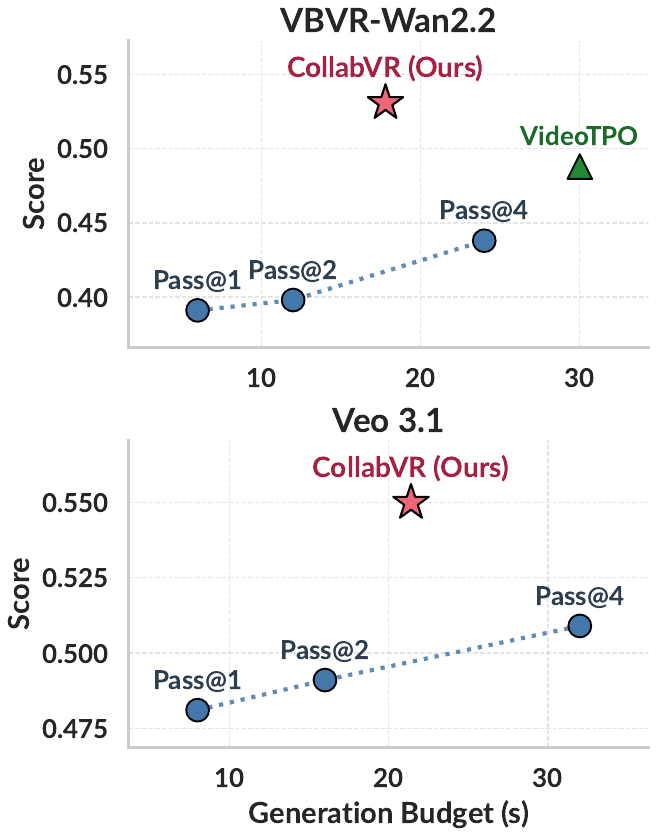}
    \vspace{-1.5em}
    \caption{
        \textbf{Performance--Cost trade-off on \genvire{}~\cite{liu2025genvire}.}
        \textcolor[HTML]{1B4D7D}{Pass@$k$ resampling} plateaus quickly with cost and \textcolor[HTML]{177829}{VideoTPO}~\cite{chen2026tivibench} trades extra budget for modest improvement, while \textcolor[HTML]{C40C22}{\oursshort{}} reaches markedly higher score at lower budget on both models~\cite{wang2026vbvr, google2026veo31}.
    }
    \vspace{-0.5em}
    \label{fig:pareto}
\end{wrapfigure}

Recent progress in visual reasoning has largely centered on the \textit{Thinking with Images} paradigm, in which Vision-Language Models (VLMs) reason through visual intermediate steps rather than purely textual ones~\cite{park2026progress, su2025thinking,hu2024sketchpad,menon2024whiteboard,li2025mvot}.
While promising, static images impose fundamental physical and cognitive limits on reasoning about dynamic processes.
A single frame captures one instant in time, and is therefore inherently constrained in representing temporal evolution, causal sequences, or physical interactions that unfold over time.

To overcome these static limitations, a new paradigm has emerged: \textit{Thinking with Video}.
Departing from symbolic, text-based Chain-of-Thought, this line of work leverages Video Generation Models (VGMs) to produce a Chain-of-Frames: a form of generative visual reasoning in which the reasoning trajectory is realized as a temporally coherent video rather than a token sequence.
dAlongside this shift, a growing suite of benchmarks now evaluates the zero-shot reasoning capabilities of VGMs~\cite{guo2025mmecof,liu2025genvire,chen2026tivibench,luo2025vreasonbench,wang2026vbvr,yang2025vrbench}.
Recent studies report that VGMs exhibit promising reasoning behavior within short sequences, covering physical dynamics, spatial consistency, fine-grained visual detail, and geometric tracing, in some cases matching or surpassing VLMs on targeted video reasoning benchmarks~\cite{wiedemer2025video,tong2026thinkingwithvideo}.

VLMs and VGMs nonetheless exhibit complementary strengths and weaknesses (Figure~\ref{fig:concept}).
VLMs excel at logical reasoning: they decompose complex problems, formulate multi-step plans, and draw abstract inferences with high fidelity, but their capacity to directly perceive and simulate the visual world remains limited.
VGMs, conversely, are strong at perceiving visual detail, preserving physical coherence, and performing short-horizon simulation, yet weak at abstract logical reasoning and long-range causal consistency.
This disparity in VGMs aligns with a broader pattern in recent evaluations: modern VGMs are trained to optimize perceptual quality, not task-level reasoning correctness~\cite{guo2025mmecof,cai2025mmgr}.

As a direct consequence, VGMs exhibit two recurring failure modes in goal-directed generative visual reasoning.
(\textit{i})~\textbf{Overloaded-prompt failure}: when a single prompt specifies a long-horizon task, the VGM collapses it into one short-horizon rollout that deviates from the intended trajectory, since it lacks the planning capacity to decompose the task into coherent sub-goals.
(\textit{ii})~\textbf{Execution failure}: even within a single short clip, the VGM commits localized errors mid-clip (e.g., an agent crossing a wall, an object losing its identity after an interaction, or a sub-action stopping before completion) that, once introduced, propagate through subsequent frames and contaminate the entire trajectory.
Both failures can be traced to a common root cause: the absence of an explicit, corrigible reasoning process, on top of the VGM's strong yet short-horizon visual prior.

In light of this, the VGM needs a reasoning supervisor that can plan beyond a single short clip and verify what was just produced, capabilities at which VLMs are already strong. A straightforward way to leverage VLMs is to select the most perceptually plausible one among $k$ samples, as in existing test-time scaling approaches~\cite{liu2025videot1,he2025evosearch,cong2025swift}.
On video reasoning tasks, however, valid outputs are task-specific and tightly defined, unlike general text-to-video which admits a wide range of realizations. More importantly, the correct trajectory often lies outside the generator's distribution and thus is difficult to reach by simply drawing more plausible samples (Figure~\ref{fig:pareto}). Therefore, in video reasoning, VLM supervision has to serve as a natural complement to the VGM’s short-horizon visual prior, most effectively when applied to progressively construct the correct trajectory through the task.

We propose \ours{} (\oursshort), a closed-loop framework that couples the VLM with the VGM at step-level granularity.
The VLM plans only the immediate next action, inspects the clip the VGM produces, and decides on the spot how to proceed, matching the intervention to the diagnosed state rather than applying one fixed recovery across whole videos.
This step-level coupling catches each failure at the moment it occurs, before it contaminates an entire trajectory.
Unlike prior work that uses a VLM as a planner or verifier, the VLM in \oursshort{} chooses each action from the frames already generated and repairs a failure at the clip where it occurs, so a fixed generator shows more of its reasoning ability at inference time without any training.
\oursshort{} consists of two tightly coupled modules, each targeting one of the failures above.

\moduleone{} addresses \textit{(i)} by letting the VLM adaptively decide the step count and plan only the immediate next action conditioned on previously generated frames, mitigating long-horizon drift without a fixed upfront decomposition; \moduletwo{} addresses \textit{(ii)} by having the VLM verify each clip, diagnose the failure, and revise the next action prompt to repair local errors before they compound.
Although the VLM is not infallible, the step-level design contains any single error to one clip, and our human-annotated reliability benchmark (Section~\ref{sec:human_benchmark}) confirms that VLM-predicted failure localization and step counts are reliable at the granularity the framework requires.

We summarize our contributions as follows.
\begin{itemize}[leftmargin=1.5em]
    \item \textbf{Progressive Planning against long-horizon drift.} An adaptive planning module where the VLM decides the step count on the fly and emits only the immediate next action, conditioned on previously generated frames.
    \item \textbf{Collaborative Reasoning against execution failure.} A failure-aware intervention module where the VLM verifies each VGM clip and folds the diagnosed failure back into the next action prompt for repair.
    \item \textbf{Consistent gains at matched compute.} \oursshort{} improves both open- and closed-source VGMs over single-inference, Pass@$k$, and VideoTPO on \genvire{} and \vbvr{}, with further gains on reasoning-fine-tuned VGMs~\cite{wang2026vbvr}. A human-annotated benchmark confirms that VLM-predicted task complexity and failure localization align with expert judgments.
\end{itemize}

\section{Related Work}
\label{sec:related_work}

\subsection{Thinking with Video: Video Reasoning}
\label{sec:rw_thinking_with_videos}

Using generated or retrieved perceptual artifacts as intermediate reasoning steps originated in the \textit{Thinking with Images} line of work on VLMs~\cite{su2025thinking,hu2024sketchpad,menon2024whiteboard,li2025mvot,su2025openthinkimg,gu2025thinkmorph, park2026progress}, where sketches, diagrams, and sub-images scaffold multi-step visual inference, but static images cannot capture dynamic processes or causal unfolding over time.
The emergence of high-fidelity Video Generation Models such as Sora~\cite{brooks2024sora}, Veo~\cite{google2026veo31}, and Wan~\cite{wanteam2025wan}, among many others~\cite{nvidia2025cosmospredict25,yang2025cogvideox,meta2024moviegen,yang2024unisim}, gives rise to the \textit{Thinking with Video} paradigm~\cite{guo2025mmecof,wiedemer2025video,tong2026thinkingwithvideo}, in which a generated video itself serves as the reasoning artifact: a Chain-of-Frames whose temporal trajectory embodies the solution.
A rapidly growing body of benchmarks~\cite{guo2025mmecof,tong2026thinkingwithvideo,luo2025vreasonbench,yang2025vrbench,liu2025genvire,chen2026tivibench,wang2026vbvr,li2026viper,cai2025mmgr} probes this regime, consistently demonstrating that modern VGMs excel at short-horizon visual simulation but remain weak at long-horizon planning, strict geometric and logical constraints, global state consistency, and process-level faithfulness, the diagnosis that motivates our framework.
Recent benchmarks and studies also separate perceptual quality from physical and causal correctness~\cite{zheng2025vbench2,wu2026worldreasonbench,kang2025howfar,xie2026yocausal} and judge the generation process in addition to the final frame~\cite{li2026viper}.
Reasoning therefore requires intermediate states that are checked and revised during generation, and \oursshort{} adds this structure at inference time.

\subsection{Test-Time Scaling for Video Generation}
\label{sec:rw_tts}

Test-time scaling (TTS) for Large Language Models~\cite{snell2024scaling,brown2024monkeys} and diffusion models~\cite{ma2025inferencetime} substantially improves output quality with additional inference compute, and video-specific extensions~\cite{liu2025videot1,he2025evosearch,cong2025swift,li2026thinkinginframes,jang2026selfrefining} apply this to the temporal axis through frame-level search, evolutionary sampling, and self-refinement.
These methods optimize \emph{visual quality} rather than task correctness, but reasoning failures are systematic (wrong solution paths, skipped sub-goals, incorrect physical outcomes) and cannot be averaged out by sampling more, requiring instead diagnosis and repair by a VLM.
Among reasoning-targeting approaches, VideoTPO~\cite{chen2026tivibench} uses LLM critique to iteratively re-prompt the generator, but its single mechanism of whole-video prompt refinement leaves task-decomposition failures unaddressed. We instead operate at sub-action granularity, with the VLM verifying each clip and folding its diagnosis directly into the next action prompt for repair.

\subsection{Iterative Refinement and VLM-Guided Generation}
\label{sec:rw_iterative}

Iterative refinement and LLM/VLM-guided generation cast an LLM as verifier or planner in a closed loop with the generator, originating in the image domain~\cite{ke2024sld,yang2024rpg} and extending to video~\cite{lin2024videodirectorgpt,yang2025vlipp,xue2025phyt2v,wang2024videoagent,huang2025vchain}, but these systems optimize visual or physical quality, treat the video as an indivisible unit, and lack a mechanism to diagnose specific failures or localize corrections.
\oursshort{} closes these gaps in a single training-free loop that plans progressively, verifies and recovers at each step based on an explicit failure diagnosis, and works with any off-the-shelf VGM.

\begin{figure}[t]
    \centering
    \includegraphics[width=\linewidth]{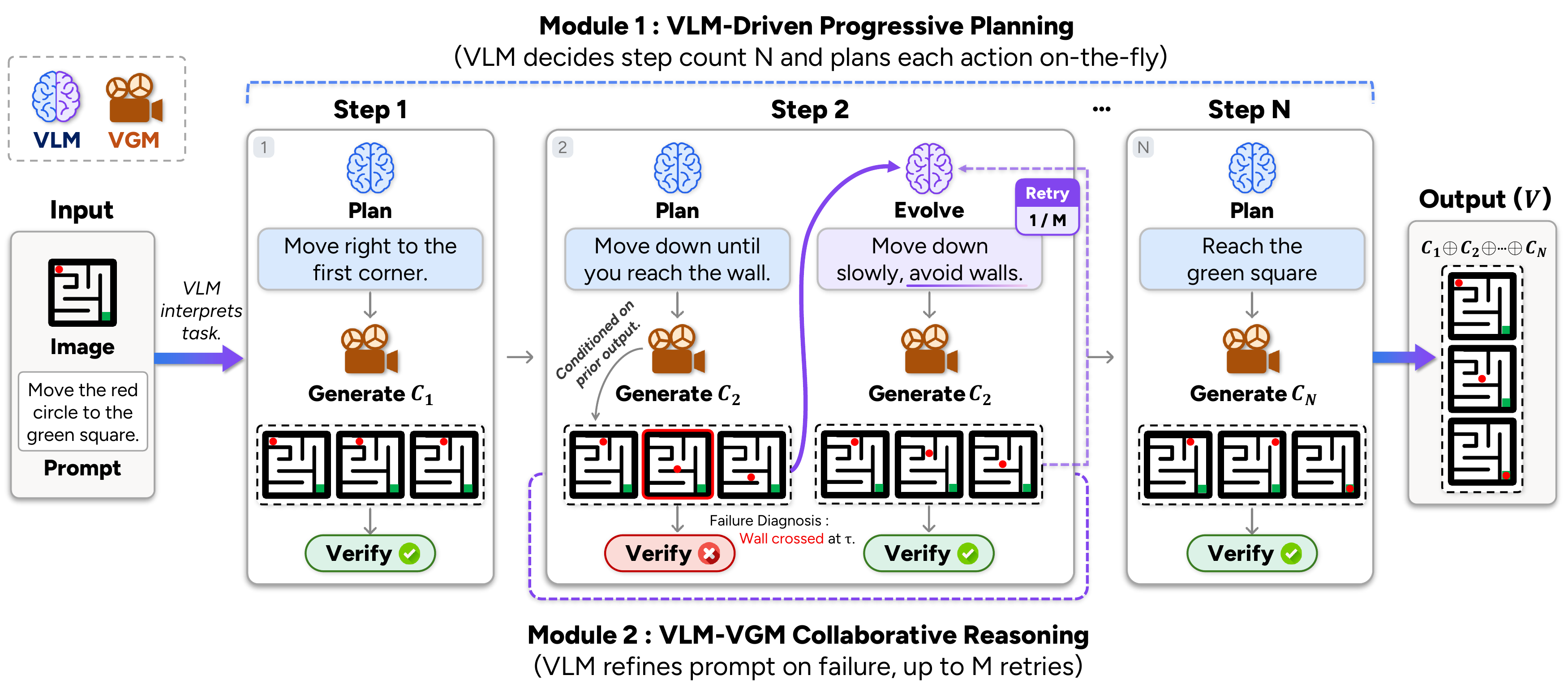}
    \caption{
        \textbf{Overall pipeline of \oursshort{}.}
        A persistent VLM plans one action at a time and, after observing each generated clip, decides whether to accept, re-generate, or re-plan. \textbf{Module~1} adaptively determines the step count, and \textbf{Module~2} verifies each clip and folds the verifier's diagnosis into the next action prompt to repair the failure.
    }
    \label{fig:pipeline}
    \vspace{-1em}
\end{figure}

\section{CollabVR: Closed-Loop Step-Level Video Reasoning}
\label{sec:method}

\begin{wrapfigure}{r}{0.5\linewidth}
\vspace{-1em}
\begin{minipage}{\linewidth}
\begin{algorithm}[H]
\footnotesize
\caption{\texttt{\oursshort} $(I_0, q, N_{\max}, M)$.}
\label{alg:collabvr}
\begin{algorithmic}[1]
\REQUIRE input image $I_0$, task prompt $q$; max planning steps $N_{\max}$; per-step attempt budget $M$
\ENSURE generated video $V$
\STATE $\mathcal{H} \gets \emptyset$,\; $f \gets I_0$
\FOR{$t = 1, \ldots, N_{\max}$}
    \STATE $a_t \gets \pi_{\text{plan}}(I_0, q, \mathcal{H})$ \COMMENT{(Section~\ref{sec:progressive_planning})}
    \FOR{$j = 1, \ldots, M$}
        \STATE $c_t \gets g(f, a_t)$
        \STATE $(v, d) \gets \pi_{\text{verify}} (I_0, q, \mathcal{H}, c_t)$ \COMMENT{(Section~\ref{sec:collab_reasoning})}
        \IF{$v = \texttt{accept}$}
            \STATE $\mathcal{H} \gets \mathcal{H} \cup \{c_t\}$
            \STATE $f \gets$ last frame of $c_t$
            \IF{task complete}
                \RETURN $V = c_1 \oplus \cdots \oplus c_t$
            \ENDIF
            \STATE \textbf{break}
        \ELSE
            \STATE $a_t \gets \mathrm{evolve}(a_t, d)$ \COMMENT{prompt evolution}
        \ENDIF
    \ENDFOR
\ENDFOR
\RETURN $V = c_1 \oplus \cdots \oplus c_{|\mathcal{H}|}$
\end{algorithmic}
\end{algorithm}
\end{minipage}
\vspace{-1em}
\end{wrapfigure}

We present \oursshort, a closed-loop test-time framework that treats video reasoning as a construction problem: the correct trajectory is assembled stepwise through VLM planning and VGM execution, rather than sampled from the generator's output distribution.
Figure~\ref{fig:pipeline} illustrates the overall pipeline, whose two core modules target the two failure modes identified in Section~\ref{sec:introduction}.
\moduleone{} (Section~\ref{sec:progressive_planning}) addresses overloaded-prompt failure by letting the VLM plan adaptively, one step at a time.
\moduletwo{} (Section~\ref{sec:collab_reasoning}) addresses execution failure by letting the VLM verify each clip and revise the action prompt with the diagnosis.

The reliability of this VLM-as-supervisor design (whether plan-depth, verification, and prompt evolution genuinely align with human judgment rather than amplifying VLM hallucinations) is empirically validated in Section~\ref{sec:analysis} on a dedicated human-annotated benchmark.

\subsection{Problem Formulation}
\label{sec:problem_formulation}

A video reasoning task is specified by an input image $I_0$ and a task prompt $q$ that describes the desired visual process or transformation, and the goal is to produce a video $V$ whose trajectory realizes the reasoning demanded by $(I_0, q)$.
Our framework has two actors: a VLM-based planner/verifier $\pi$ (queried in two roles, $\pi_{\text{plan}}$ and $\pi_{\text{verify}}$) and an image-to-video generator $g$ that maps a conditioning frame $f$ and an action prompt $a_t$ to a short clip $c_t$.
Throughout the loop we maintain $f$ as the latest conditioning frame (initially $I_0$) and $\mathcal{H}$ as the history of accepted clips.
The full closed loop is given in Algorithm~\ref{alg:collabvr}, and the output is the concatenation of accepted clips $V = c_1 \oplus \cdots \oplus c_N$; the rest of this section motivates its two design choices, progressive planning (Section~\ref{sec:progressive_planning}) and collaborative reasoning (Section~\ref{sec:collab_reasoning}).

\subsection{VLM-Driven Progressive Planning}
\label{sec:progressive_planning}

\begin{figure}[t]
    \centering
    \begin{minipage}[b]{0.64\linewidth}
        \centering
        \includegraphics[width=\linewidth]{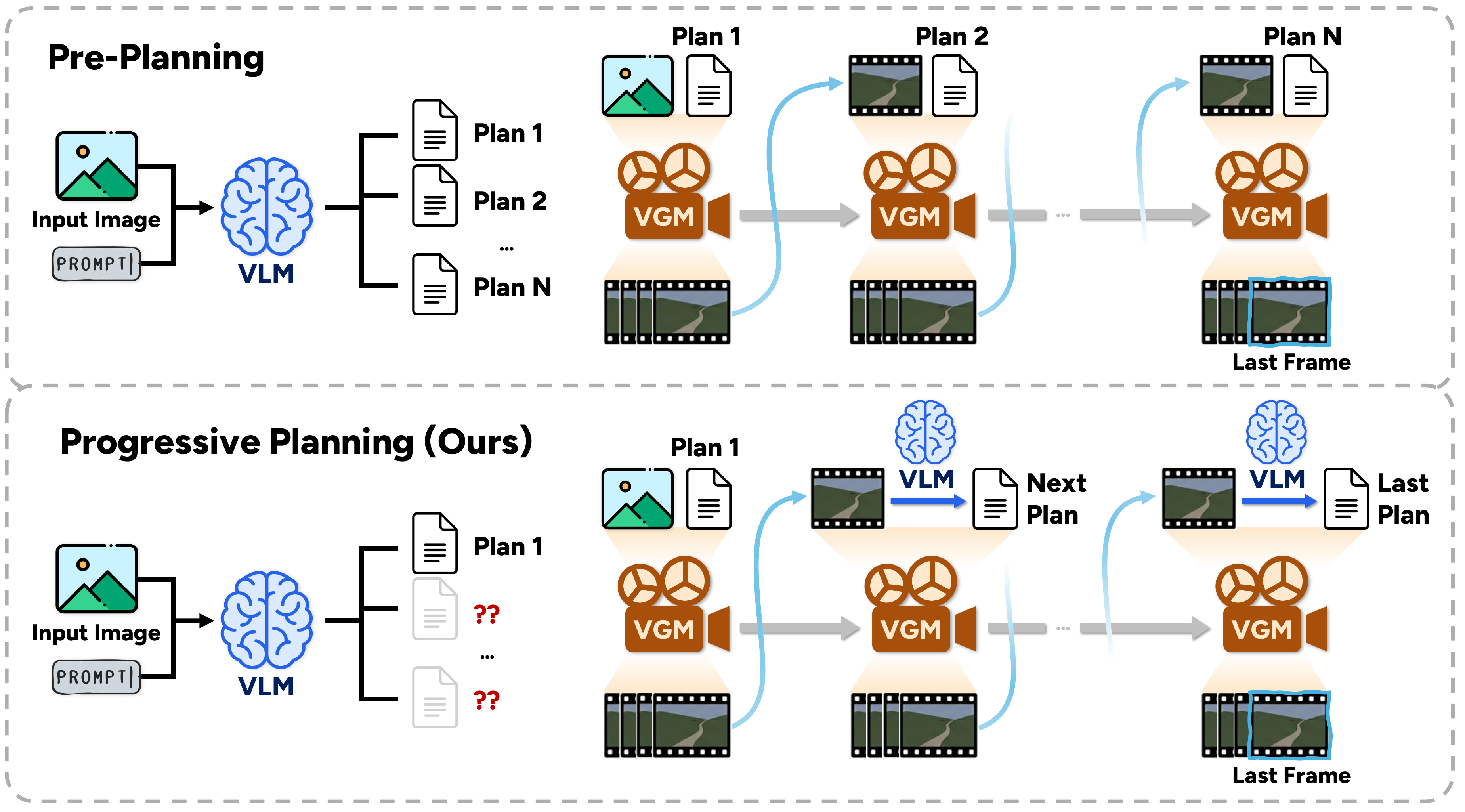}
        \subcaption{Pre-planning vs.\ Progressive planning pipeline.}
        \label{fig:preplan_vs_progplan_pipeline}
    \end{minipage}
    \hfill
    \begin{minipage}[b]{0.34\linewidth}
        \centering
        \includegraphics[width=\linewidth]{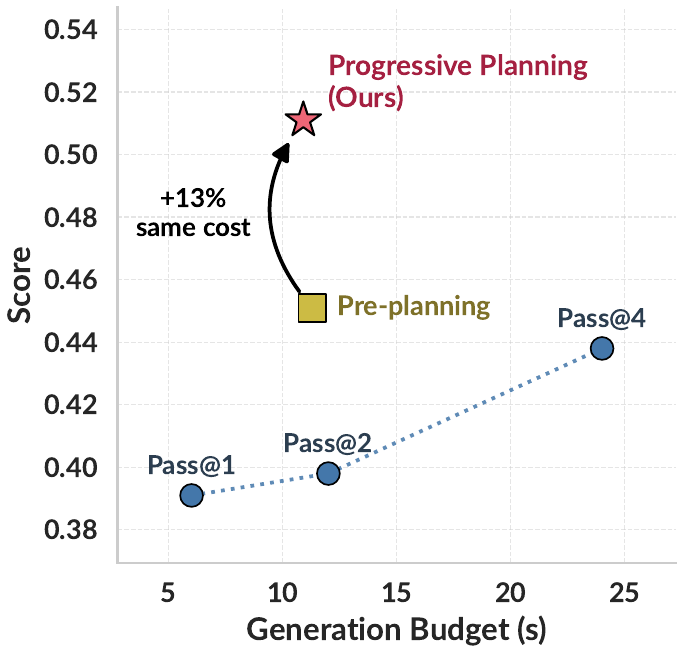}
        \subcaption{Performance--cost on \genvire{}.}
        \label{fig:preplan_vs_progplan_pareto}
    \end{minipage}
    \caption{
        \textbf{Pre-planning vs.\ Progressive planning} on \genvire{} with \vbvrwan{} (Module~1 only). Progressive planning achieves a $+13\%$ relative gain over pre-planning at matched cost.
    }
    \vspace{-1em}
    \label{fig:progressive_planning}
\end{figure}

A naive extension of Chain-of-Thought to video is \emph{pre-planning}~\cite{huang2025vchain} (Figure~\ref{fig:progressive_planning}a, top): the VLM decomposes $q$ into $N$ milestone prompts upfront, and the VGM sequentially generates a clip per milestone.
This reduces drift on long-horizon tasks but commits the plan before any VGM output exists, so it cannot adapt to the realized generation, and $N$ itself is difficult to fix from the prompt alone.
We therefore adopt \emph{progressive planning} (Figure~\ref{fig:progressive_planning}a, bottom; line 4 of Algorithm~\ref{alg:collabvr}): the VLM plans only the immediate next action and inspects the realized clip before deciding whether to continue, so both subsequent steps and $N$ adapt to what the generator actually produces, with $N$ capped at a hyperparameter $N_{\max}$.
At matched generation budgets, this yields a substantially better performance--cost trade-off than pre-planning (Figure~\ref{fig:progressive_planning}b).

\subsection{VLM-VGM Collaborative Reasoning}
\label{sec:collab_reasoning}

For each generated clip $c_t$, the VLM verifier $\pi_{\text{verify}}$ produces a structured judgment $(v, d)$, where $v \in \{\text{accept}, \text{reject}\}$ and $d$ packages a textual reason and an actionable suggestion for repair (line 6 of Algorithm~\ref{alg:collabvr}).
The verifier judges whether the planned action was executed, flagging wrong direction, wrong target, or scene collapse but not partial-but-correct progress, which is the planner's concern.
This single-clip judgment stays within VLMs' strong perception regime, so $d$ is specific enough to be actionable.
The simplest recovery folds $d$'s suggestion (e.g., \textit{``circle the red square, not the yellow''}) back into the action prompt, $a_t \gets \mathrm{evolve}(a_t, d)$, and re-samples the VGM with the plan fixed, reusing the verifier's output without an extra VLM call.

\section{Experiments}
\label{sec:experiments}

We evaluate \oursshort{} on \genvire{} and \vbvr{}, two complementary video reasoning benchmarks.
The section presents the experimental setup (Section~\ref{sec:implementation}), main accuracy comparisons under matched compute (Section~\ref{sec:main_results}), per-module ablations (Section~\ref{sec:ablation}), and an analysis covering per-category module effectiveness and the human-annotated reliability of the VLM-as-supervisor design (Section~\ref{sec:analysis}).
Complementary evidence in the appendix includes a blind user study (Appendix~\ref{sec:appendix_user_study}) and a cost decomposition validating the matched-compute framing (Appendix~\ref{sec:appendix_cost}).

\begin{table*}[t]
\centering
\scriptsize
\captionsetup{font=small}
\setlength{\tabcolsep}{4pt}
\caption{Benchmarking results on \genvire{}. All scores are 3-run VLM evaluation averages (\gemini{}) for stability. \textbf{VGM Cost} is the average total VGM generation seconds per sample (all steps and re-generations included). Higher is better. \textbf{Bold}: best within each model; \underline{Underlined}: second best.}
\resizebox{0.8\linewidth}{!}{
\begin{tabular}{l|c|c|cccccc}
\toprule
 & & & \multicolumn{6}{c}{\textbf{Category}} \\
\cmidrule(lr){4-9}
\textbf{Method} & \smash{\makecell[b]{\textbf{VGM} \\ \textbf{Cost (s)}}} & \textbf{Avg.} & \textbf{Abst.} & \textbf{Algo.} & \textbf{Analog.} & \textbf{Perc.} & \textbf{Plan.} & \textbf{Spat.} \\
\midrule

\rowcolor{gray!10}\textbf{Open-source Video Models} & & & & & & & & \\

\vbvrwan & 6.0 & 0.391 & 0.479 & 0.415 & 0.250 & 0.261 & 0.554 & 0.387 \\
\vbvrwan{} + Pass@2 & 12.0 & 0.398 & \underline{0.576} & 0.437 & 0.278 & 0.257 & 0.481 & 0.357 \\
\vbvrwan{} + Pass@4 & 24.0 & 0.438 & \textbf{0.622} & 0.418 & 0.250 & 0.275 & 0.604 & 0.462 \\
\vbvrwan{} + VideoTPO~\cite{chen2026tivibench} & 30.0 & \underline{0.488} & 0.535 & \underline{0.443} & \textbf{0.417} & \underline{0.313} & \underline{0.671} & \textbf{0.552} \\
\rowcolor{line-blue}
\vbvrwan{} + \textbf{\oursshort{}} & 17.8 & \textbf{0.531} & 0.569 & \textbf{0.606} & \underline{0.333} & \textbf{0.367} & \textbf{0.821} & \underline{0.488} \\

\midrule

\rowcolor{gray!10}\textbf{Closed-source Video Models} & & & & & & & & \\

\veo & 8.0 & 0.481 & 0.420 & 0.512 & 0.361 & 0.274 & \underline{0.744} & 0.573 \\
\veo{} + Pass@$2$ & 16.0 & 0.491 & \textbf{0.458} & \underline{0.587} & 0.389 & 0.242 & 0.721 & 0.571 \\
\veo{} + Pass@$4$ & 32.0 & \underline{0.509} & \underline{0.425} & 0.573 & \underline{0.417} & \underline{0.296} & 0.726 & \underline{0.646} \\
\rowcolor{line-blue}
\veo{} + \textbf{\oursshort{}} & 21.4 & \textbf{0.550} & 0.434 & \textbf{0.641} & \textbf{0.472} & \textbf{0.325} & \textbf{0.768} & \textbf{0.657} \\

\bottomrule
\end{tabular}
}
\label{tab:genvire_results}
\vspace{-0em}
\end{table*}

\begin{table*}[t]
\vspace{-0.5em}
\centering
\scriptsize
\captionsetup{font=small}
\setlength{\tabcolsep}{3pt}
\caption{Benchmarking results on \vbvr{}. Overall In-Domain (ID) and Out-of-Domain (OOD) scores are reported alongside category-wise performance. Higher is better. \textbf{Bold}: best in group; \underline{Underlined}: second best.}
\resizebox{1.0\linewidth}{!}{
\begin{tabular}{l|c|c|c|ccccc|c|ccccc}
\toprule
& \multicolumn{1}{c|}{\textbf{}}
& \multicolumn{1}{c|}{\textbf{}}
& \multicolumn{6}{c|}{\textbf{In-Domain by Category}}
& \multicolumn{6}{c}{\textbf{Out-of-Domain by Category}} \\
\cmidrule(lr){4-9}
\cmidrule(lr){10-15}
\textbf{Models}
& \smash{\makecell[b]{\textbf{VGM} \\ \textbf{Cost (s)}}}
& \textbf{Overall}
& \textbf{Avg.}
& \textbf{Abst.} & \textbf{Know.} & \textbf{Perc.} & \textbf{Spat.} & \textbf{Trans.}
& \textbf{Avg.}
& \textbf{Abst.} & \textbf{Know.} & \textbf{Perc.} & \textbf{Spat.} & \textbf{Trans.} \\
\midrule

\vbvrwan{} & 3.70 & 0.671 & 0.762 & 0.701 & \underline{0.746} & 0.802 & 0.793 & 0.803 & 0.577 & 0.674 & \textbf{0.674} & 0.503 & 0.528 & 0.633 \\

\vbvrwan{} + Pass@$2$ & 7.40 & 0.694 & 0.783 & \underline{0.791} & 0.742 & 0.795 & 0.774 & 0.812 & 0.602 & 0.728 & 0.617 & 0.494 & 0.532 & \underline{0.701} \\

\vbvrwan{} + Pass@$4$ & 14.80 & \underline{0.707} & \underline{0.789} & 0.751 & 0.734 & \textbf{0.826} & \underline{0.805} & \underline{0.841} & \underline{0.622} & \underline{0.785} & \underline{0.660} & \underline{0.535} & \underline{0.577} & 0.683 \\

\vbvrwan{} + VideoTPO~\cite{chen2026tivibench} & 11.10 & 0.650 & 0.717 & 0.723 & 0.698 & 0.641 & 0.744 & 0.816 & 0.582 & 0.767 & 0.619 & 0.513 & 0.540 & 0.572 \\

\rowcolor{line-blue} \vbvrwan{} + \textbf{\oursshort{}} & 10.91 & \textbf{0.757} & \textbf{0.819} & \textbf{0.828} & \textbf{0.784} & \underline{0.805} & \textbf{0.828} & \textbf{0.852} & \textbf{0.696} & \textbf{0.884} & 0.634 & \textbf{0.641} & \textbf{0.608} & \textbf{0.720} \\
\cdashline{1-15}

\rule{-0.5pt}{8.0pt}Cosmos-Predict2.5 & 3.70 & 0.308 & 0.312 & 0.272 & 0.327 & 0.355 & 0.227 & 0.390 & 0.304 & 0.368 & 0.169 & 0.309  & 0.377 & 0.274 \\
\rowcolor{line-blue} Cosmos-Predict2.5 + {\oursshort{}}
& 10.91
& \textbf{0.403}
& \textbf{0.406}
& \textbf{0.404}
& \textbf{0.431}
& \textbf{0.411}
& \textbf{0.301}
& \textbf{0.482}
& \textbf{0.400}
& \textbf{0.481}
& \textbf{0.286}
& \textbf{0.400}
& \textbf{0.471}
& \textbf{0.346} \\

\bottomrule
\end{tabular}
}
\label{tab:vbvr_results}
\end{table*}

\subsection{Implementation Details}
\label{sec:implementation}

\paragraph{Benchmarks and evaluation.}
We evaluate \oursshort{} on two complementary video-reasoning benchmarks.
\textbf{\genvire}~\cite{liu2025genvire} contains $72$ samples across 6 categories and uses a rubric-based VLM judge (\gemini{}) that scores each generated video against per-task criteria targeting task correctness rather than visual quality.\footnote{We report \genvire{} scores averaged over three runs to account for judge stochasticity.}
\textbf{\vbvr{}}~\cite{wang2026vbvr} focuses on synthetic visual reasoning tasks with controlled ground-truth references. It consists of a $500$-sample test set spanning 5 reasoning categories with In-Domain and Out-of-Domain splits, and adopts a rule-based, judge-free protocol for comparing generated videos against ground-truth references.

\paragraph{Video generation models.}
We apply \oursshort{} to two main VGMs: the closed-source API model\footnote{We exclude \sora{} from the main results due to insufficient first-frame fidelity for step-by-step clip concatenation, and defer its analysis to Appendix~\ref{sec:appendix_sora}.} \textbf{\veo}~\cite{google2026veo31}, and \textbf{\vbvrwan}~\cite{wanteam2025wan,wang2026vbvr}, a $14$B open-source image-to-video model that is the strongest open baseline on \vbvr{}.
We additionally report on \textbf{Cosmos-Predict-2.5}~\cite{nvidia2025cosmospredict25} as a second open-source VGM.

\paragraph{Baselines.}
We compare \oursshort{} against three baselines: (i) \emph{Single Inference}, one-shot generation from $(I_0, q)$; (ii) \emph{Pass@$k$}, $k \in \{2, 4\}$ independent generations from which a VLM selects the one that best achieves the task; and (iii) \emph{VideoTPO}~\cite{chen2026tivibench}, a TTS-for-video-reasoning baseline that iteratively rewrites prompts based on full-video critiques. Both (ii) and (iii) use \gemini{}, the same model used as our step-level verifier; full configuration is in Appendix~\ref{sec:appendix_hyperparams}.
Since VLM compute is negligible relative to VGM compute, we report Cost throughout as the total seconds of video generated by the VGM per sample.

\paragraph{Our framework.}
We use \gemini{} as the default VLM for both planning and verification, and evaluate alternative VLMs in Section~\ref{sec:ablation}.
Unless stated otherwise, we set $N_{\max}{=}3$ and the per-step attempt budget to $M{=}3$; the resulting total generation budget is comparable to Pass@$2$--Pass@$4$ for a single VGM.

\begin{figure*}[t]
    \centering
    \includegraphics[
        width=0.95\textwidth,
        height=\textheight,
        keepaspectratio
    ]{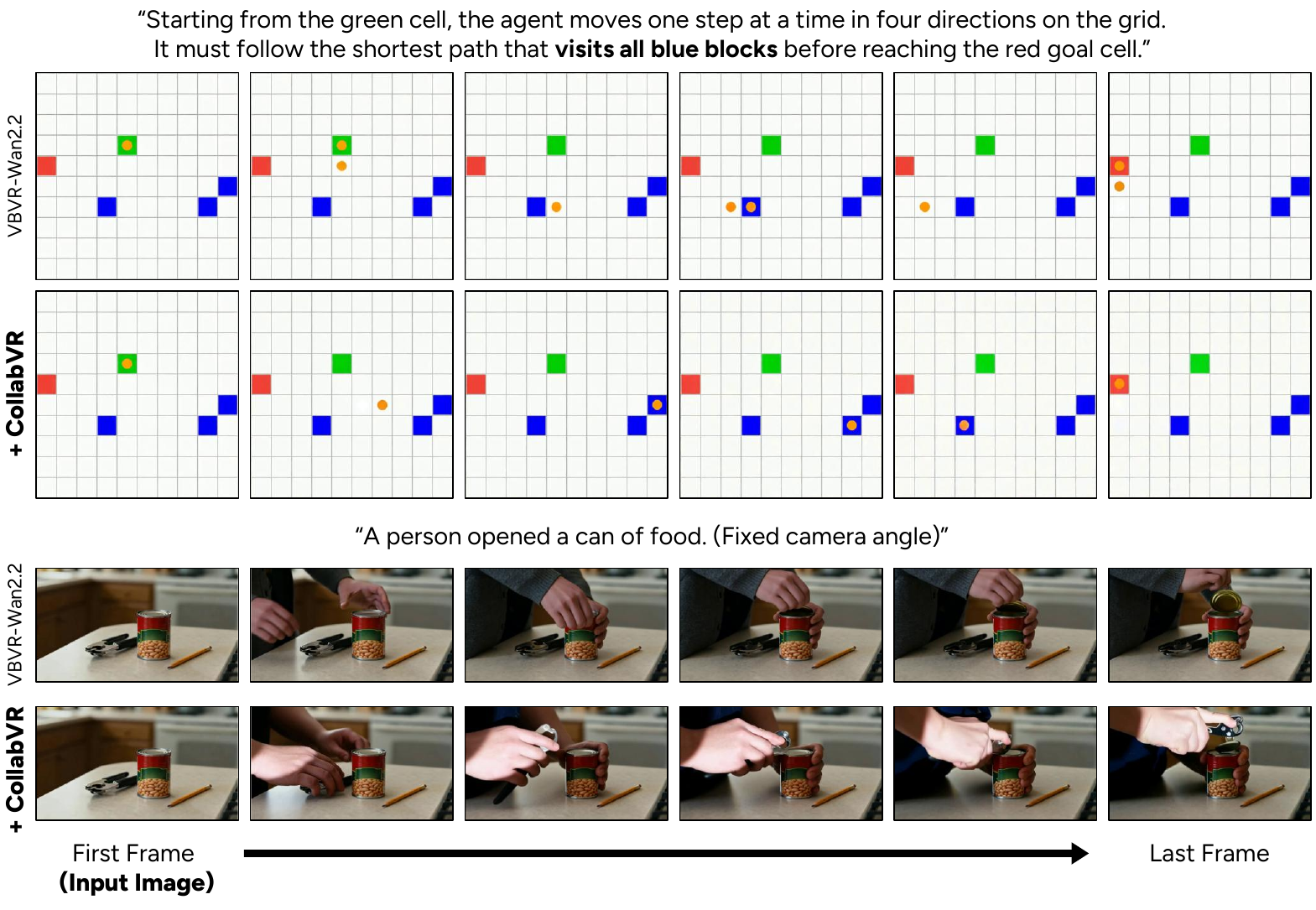}
    \caption{
        Qualitative comparison on various visual reasoning tasks from \genvire{} and \vbvr{}. We compare \vbvrwan{} with \oursshort{} on diverse tasks that require step-by-step reasoning.
    }
    \label{fig:main_visualization}
    \vspace{-0.5em}
\end{figure*}

\subsection{Main Results}
\label{sec:main_results}

Tables~\ref{tab:genvire_results} and~\ref{tab:vbvr_results} report our main results on \genvire{} and \vbvr{}.
On \genvire{}, \oursshort{} delivers consistent improvements over single-inference on both the open-source \vbvrwan{} (Pass@1~$0.391 \rightarrow$~\textbf{0.531}) and the closed-source \veo{} (Pass@1~$0.481 \rightarrow$~\textbf{0.550}) (Table~\ref{tab:genvire_results}), with the largest margins over baselines on Planning and Algorithmic categories where complex long-horizon reasoning is required.

\begin{wrapfigure}{R}{0.4\linewidth}
\vspace{-0.5em}
\begin{minipage}{\linewidth}
\centering
\captionof{table}{Per-module ablation on \genvire{} and \vbvr{}. $\Delta$: gain over the baseline (no module).}
\setlength{\tabcolsep}{7pt}
\label{tab:method_ablation}
\scriptsize
\resizebox{\linewidth}{!}{%
            \begin{tabular}{ccccc}
                \Xhline{3\arrayrulewidth}
                \rule{-2.5pt}{10.0pt}
                \bf M1 & \bf M2
                & \bf Cost (s)
                & \bf Overall
                & \bf $\Delta$ \\
                \midrule
                \rowcolor{line-blue}\multicolumn{5}{c}{\textbf{\genvire{}}} \\
                \textcolor[HTML]{A30000}{\xmark} & \textcolor[HTML]{A30000}{\xmark} & 6.0  & 0.391 & -- \\
                \textcolor[HTML]{007304}{\cmark} & \textcolor[HTML]{A30000}{\xmark} & 10.9 & 0.511 & +0.120 \\
                \textcolor[HTML]{A30000}{\xmark} & \textcolor[HTML]{007304}{\cmark} & 9.9  & 0.436 & +0.045 \\
                \textcolor[HTML]{007304}{\cmark} & \textcolor[HTML]{007304}{\cmark} & 17.8 & \bf 0.531 & \bf +0.140 \\
                \midrule
                \rowcolor{line-blue}\multicolumn{5}{c}{\textbf{\vbvr{}}} \\
                \textcolor[HTML]{A30000}{\xmark} & \textcolor[HTML]{A30000}{\xmark} & 3.70  & 0.671 & -- \\
                \textcolor[HTML]{007304}{\cmark} & \textcolor[HTML]{A30000}{\xmark} & 6.19  & 0.706 & +0.035 \\
                \textcolor[HTML]{A30000}{\xmark} & \textcolor[HTML]{007304}{\cmark} & 6.03  & 0.734 & +0.063 \\
                \textcolor[HTML]{007304}{\cmark} & \textcolor[HTML]{007304}{\cmark} & 10.91 & \bf 0.757 & \bf +0.086 \\
                \Xhline{3\arrayrulewidth}
            \end{tabular}%
}

\vspace{0.6em}

\includegraphics[width=0.85\linewidth]{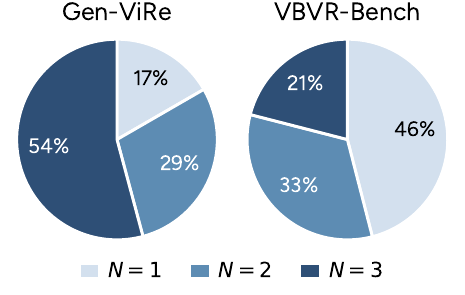}
\vspace{-1em}
\captionof{figure}{Human-annotated distribution of step counts $N$ for the benchmarks.}
\label{fig:n_distribution}
\end{minipage}
\vspace{-2em}
\end{wrapfigure}

On \vbvr{}, \oursshort{} consistently surpasses baselines on both open-source models, \vbvrwan{} and Cosmos-Predict-2.5 (Table~\ref{tab:vbvr_results}), with gains most pronounced on categories requiring multi-step spatial or transformation reasoning.
Compared to VideoTPO and Pass@$k$, \oursshort{} achieves higher accuracy at lower per-sample generation cost, supporting our claim that adaptive progressive planning coupled with failure-aware recovery is a more effective test-time scaling axis than full-video resampling.
Notably, \oursshort{} also yields further gains on top of \vbvrwan{}, a VGM already fine-tuned on reasoning data, demonstrating that test-time reasoning supervision is orthogonal to and stacks on reasoning-oriented fine-tuning.

\paragraph{Qualitative results.}

Figure~\ref{fig:main_visualization} shows representative examples where \oursshort{} contributes over the baseline.
On a long-horizon task (top), \vbvrwan{} alone cannot solve in a single shot, whereas \oursshort{} succeeds by decomposing the plan to visit the goal cells starting from the right. The framework also extends to real-world scenarios (bottom): for \textit{``opened a can of food''}, \vbvrwan{} alone bypasses the can opener and pries the lid by hand, whereas \oursshort{}'s planner emits explicit tool-use sub-actions and the VGM produces a faithful execution. Beyond automated metrics, we further confirm via a user study that human annotators prefer \oursshort{}'s outputs ($73.8\%$) over Pass@$4$ ($19.7\%$) and Pass@$1$ ($6.5\%$) on a blind side-by-side comparison (Appendix~\ref{sec:appendix_user_study}).

\begin{figure}[t]
    \centering
    \includegraphics[
    height=0.7\textheight,
    width=\linewidth,
    keepaspectratio
]{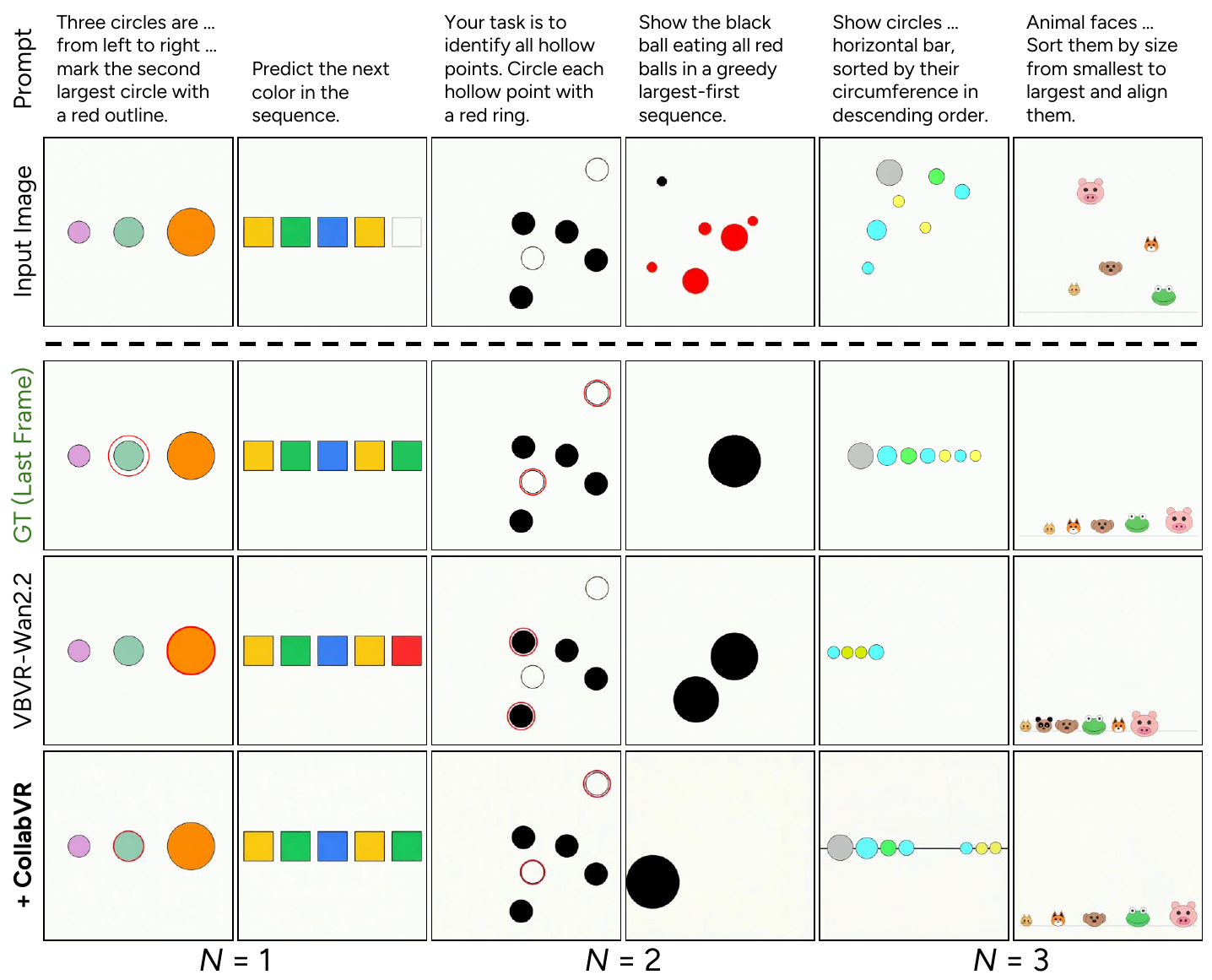}
\caption{
\textbf{Qualitative gains scale with planner-predicted step count $N$ on \vbvr{}.}
For each task we show the input image, GT last frame, single-shot \vbvrwan{} output, and \vbvrwan{}+\oursshort{} output, with two representative tasks grouped under each of $N{=}1$, $2$, and $3$.
}
    \label{fig:VBVR_VBVRWan2.2_by_N}
\end{figure}

Figure~\ref{fig:VBVR_VBVRWan2.2_by_N} groups \vbvr{} examples by the planner's predicted step count $N$.
For atomic tasks ($N{=}1$), both methods produce similar outputs and \oursshort{} only removes incidental errors such as a wrong next color.
As $N$ grows, single-shot failures become structural.
On \textit{``sort them by size from smallest to largest and align them''}, \vbvrwan{} substitutes and misorders the animal faces, whereas \oursshort{} reproduces the GT arrangement.

\subsection{Ablation Study}
\label{sec:ablation}

\paragraph{Module 1 vs.\ Module 2 vs.\ Combined.}
We compare three configurations: (M1) progressive planning only (no verification or recovery), (M2) verification and failure-aware recovery only ($N_{\max}{=}1$, no progressive planning), and (M1+M2) the full pipeline.
Figure~\ref{fig:n_distribution} shows the human-annotated distribution of step counts $N$ for the two benchmarks; this distribution reflects each benchmark's mix of task complexity and type, with \genvire{} dominated by multi-step tasks and \vbvr{} dominated by single-step tasks.
Table~\ref{tab:method_ablation} reports the corresponding per-module gains: on \genvire{}, M1's progressive planning is the larger contributor ($+0.120$ vs.\ $+0.045$ for M2), whereas on \vbvr{}, M2 is the larger contributor ($+0.063$ vs.\ $+0.035$ for M1).
On \genvire{}, \oursshort{} frequently improves performance by adaptively decomposing complex tasks into sub-steps the VGM can satisfy individually.
On \vbvr{}, M2 corrects single-clip execution failures without unnecessarily splitting tasks that are already solvable in a single action.
That the dominant module shifts with the benchmark's $N$ profile indicates the framework adapts to task character rather than relying on a single fixed mechanism.
M1 also presupposes M2.
Without verification, each planned step has to succeed in one shot as written.
The reasoning-tuned \vbvrwan{} manages this but \veo{} does not, so M1 alone falls below single inference on \veo{} ($0.481 \to 0.446$) while the full pipeline reaches $0.550$ (Appendix~\ref{sec:appendix_veo_ablation}).

\begin{figure*}[t]
    \centering
    \vspace{-1em}
    \begin{minipage}[t]{0.37\linewidth}
        \vspace{0pt}
        \centering
        \includegraphics[width=\linewidth]{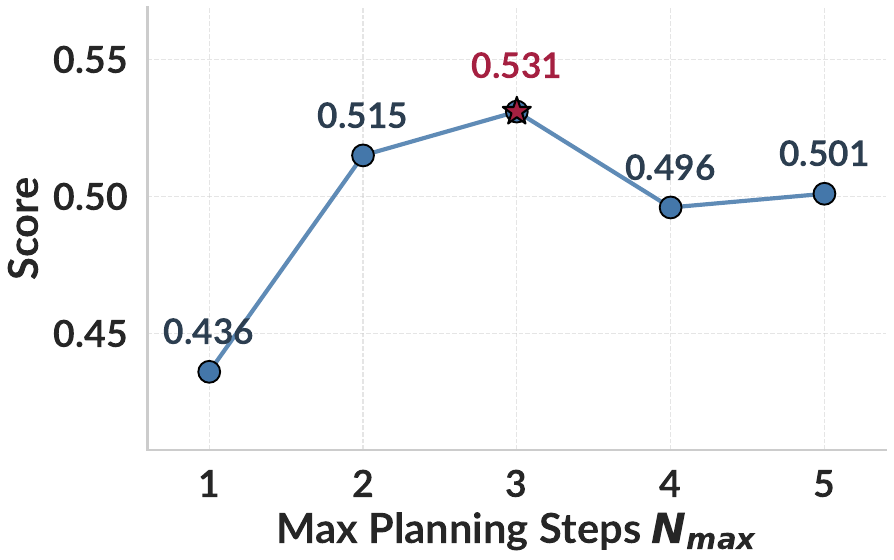}
        \captionof{figure}{
            Effect of maximum planning steps $N_{\max}$ on \genvire{}.
        }
        \label{fig:n_ablation}
    \end{minipage}
    \hfill
    \begin{minipage}[t]{0.50\linewidth}
    \vspace{0pt}
    \centering
    \scriptsize
    \setlength{\tabcolsep}{1pt}
    \captionof{table}{
        Comparison of test-time scaling methods and VLM choices across benchmarks.
    }
    \resizebox{\linewidth}{!}{
        \begin{tabular}{llcc}
            \Xhline{3\arrayrulewidth}
            \rule{-2.5pt}{10.0pt}
            \textbf{Method} & \textbf{VLM} & \textbf{\genvire{}} & \textbf{\vbvr{}} \\
            \midrule
            Pass@$1$ (baseline)
            & \multicolumn{1}{c}{--}
            & 0.391
            & 0.671 \\
            Pass@$2$
            & \gemini{}
            & 0.398
            & 0.694 \\
            Pass@$4$
            & \gemini{}
            & 0.438
            & 0.707 \\
            VideoTPO~\cite{chen2026tivibench}
            & \gemini{}
            & 0.488
            & 0.650 \\
            \midrule
            \multirow{3}{*}{\oursshort{}}
            & Qwen3.5-9B
            & 0.514
            & 0.710 \\
            & Qwen3.5-27B
            & 0.510
            & 0.717 \\
            & \gemini{}
            & \textbf{0.531}
            & \textbf{0.757} \\
            \Xhline{3\arrayrulewidth}
        \end{tabular}
    }
    \label{tab:tts_vlm_comparison}
    \end{minipage}
    \vspace{-1em}
\end{figure*}

\paragraph{Effect of step count $N_{\max}$.}
We sweep the planning step count $N_{\max} \in \{1, \ldots, 5\}$ on \genvire{} with \vbvrwan{} and observe a non-monotonic relationship: scores rise as $N_{\max}$ increases up to the level required by the task, then plateau or degrade as further splitting introduces step-boundary artifacts on already-simple sub-actions.
This empirically motivates the adaptive $N$ selection in Section~\ref{sec:progressive_planning}.

\paragraph{VLM choice.}
We replace the default \gemini{} planner/verifier with open-source alternatives (Qwen3.5-27B, Qwen3.5-9B~\cite{qwen2026qwen35}) and report the final task accuracy on each benchmark (Table~\ref{tab:tts_vlm_comparison}).
Performance degrades gracefully with weaker VLMs, and notably even the smallest model we test, Qwen3.5-9B paired with \oursshort{}, surpasses every Pass@$k$ and VideoTPO baseline that uses the proprietary \gemini{} on both benchmarks, confirming that the framework is not tied to a single proprietary model.
We further examine in Section~\ref{sec:analysis} how these end-to-end gaps are consistent with each VLM's per-axis supervisor quality (plan-depth, verification, evolution) on the human-annotated benchmark.

\subsection{Analysis}
\label{sec:analysis}

\paragraph{Category-wise module effectiveness and limitation.}
\label{sec:cat_module}

\begin{wrapfigure}{R}{0.5\linewidth}
    \vspace{-3em}
    \centering
    \includegraphics[width=\linewidth]{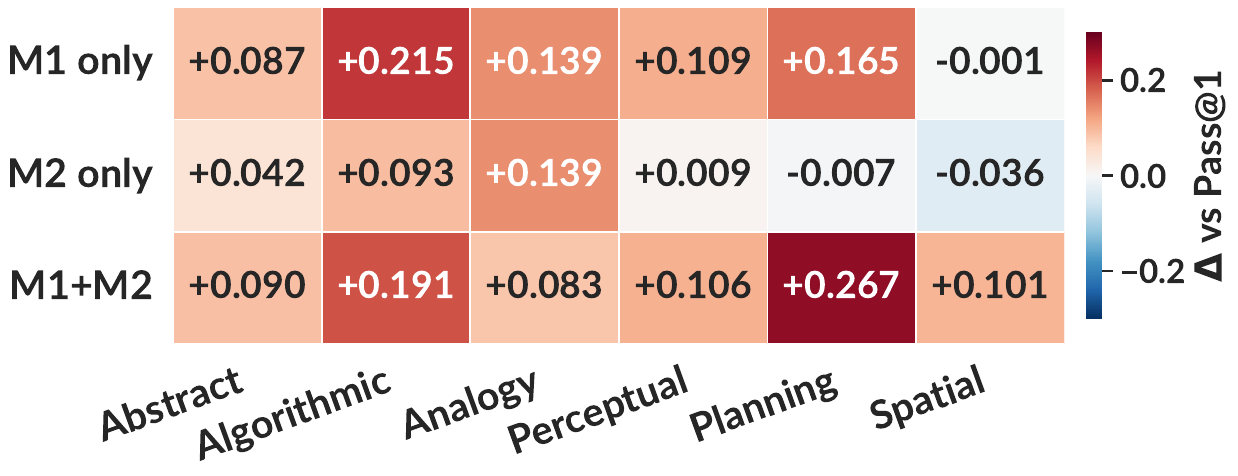}
    \caption{
\textbf{Per-category $\Delta$ over Pass@1 on \genvire{} (\vbvrwan{}).}
Module configurations follow Section~\ref{sec:ablation}.
}
    \label{fig:category_module}
    \vspace{-0em}
\end{wrapfigure}

CollabVR does not behave uniformly across reasoning types.
Section~\ref{sec:ablation} attributed module dominance to a benchmark's $N$ profile.
Here we examine the finer-grained role of each reasoning category (Figure~\ref{fig:category_module}).
For instance, a Planning prompt such as \textit{``use the kettle and teabag to make a cup of tea in the mug''} packs a chain of distinct physical actions into a deceptively short instruction; a single-shot VGM compresses these into one ambiguous clip and typically resolves only the first action, whereas the progressive planner exposes the chain as separate sub-goals that the VGM can satisfy individually (M1 alone $+0.165$).
Conversely, an Analogy prompt such as \textit{``generate the missing object in the lower right region and solve the visual analogy''} is a single atomic transformation: verifier-driven re-sampling alone is sufficient (M2 alone $+0.139$).
Crucially, the full M1+M2 pipeline yields a positive gain in every category ($+0.083$ to $+0.267$), and only the combination meaningfully improves long-horizon Spatial tasks, indicating decomposition and recovery are complementary

Yet the framework cannot manufacture capabilities the VGM lacks: the smallest gains concentrate on categories whose target transformations are symbolic rather than physical (Analogy $\Delta\,{+}0.083$, Abstract $\Delta\,{+}0.090$), where decomposition has nothing to decompose into and verifier-driven re-sampling can only redraw from the VGM's existing distribution.
On Analogy in particular, M1+M2 even underperforms either module alone (both $+0.139$), since forcing decomposition on an atomic transformation yields contrived intermediates that the verifier then rejects.
This residual gap is orthogonal to test-time orchestration, and we view reasoning-oriented VGM training (e.g., physics-aware fine-tuning, symbolic-transformation pretraining) as a complementary future direction.

\begin{figure}[t]
    \centering
    \includegraphics[width=\linewidth]{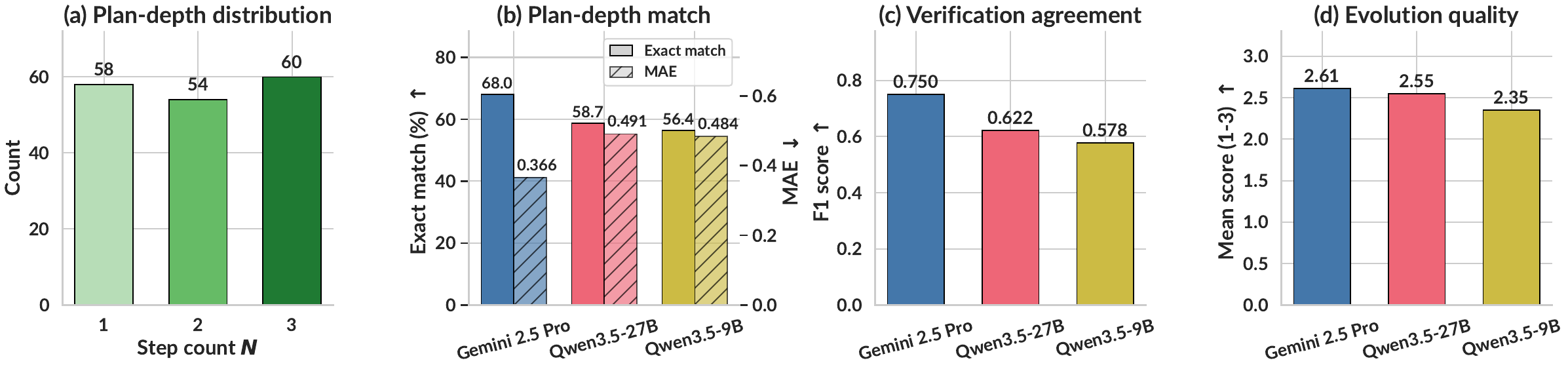}
    \caption{
\textbf{Human-annotated analysis of planning, verification, and evolution.}
(a) Distribution of human-annotated step counts $N$.
(b) Plan-depth match per VLM: exact-match accuracy (left axis) and mean absolute error (MAE) (right axis).
(c) Verification agreement: F1 score on a balanced 1:1 split.
(d) Evolution quality: mean human rating on a three-point scale.
}
    \label{fig:human_eval}
    \vspace{-1em}
\end{figure}

\paragraph{Human-annotated benchmark and VLM supervision.}
\label{sec:human_benchmark}

To externally check the VLM-as-supervisor assumption underlying the closed loop, we construct a human-annotated benchmark on \vbvrwan{} output videos, with each item annotated along three axes that mirror the framework's decision points.
\textit{(i) Plan depth.} From the input image and prompt alone, annotators decide the appropriate step count $N$ for solving the task, which reflects the task's reasoning complexity, compared against the VLM's adaptive $N$ from Module~1.
\textit{(ii) Clip-level verification.} For 250 clips drawn at the verifier's decision points, annotators independently judge whether the clip realizes the intended sub-action; this is compared against the VLM verifier's accept/reject output from Module~2.
\textit{(iii) Evolution scoring.} For 240 rejected clips, annotators rate the suitability of the verifier's suggested repair (used by prompt evolution) on a three-point scale (1: poor, 2: moderate, 3: well-suited).

For reliability and fairness, the accept/reject ratio is held to 1:1 (preventing class-imbalance inflation of verifier accuracy) and the $N$ distribution is balanced across reasoning categories.
Each item is independently annotated by AI experts from two different affiliations, with disagreements resolved through cross-validation, providing a non-trivial inter-annotator baseline against which VLM agreement can be calibrated.

With \gemini{}~\cite{geminiteam2025gemini25} as the planner/verifier, the VLM aligns most closely with human annotators on all three axes (Figure~\ref{fig:human_eval}), supporting its role as the default planner/verifier in our pipeline.
The same axes are used in the VLM ablation (Section~\ref{sec:ablation}) to compare alternative open-source VLMs (Qwen3.5-27B, Qwen3.5-9B~\cite{qwen2026qwen35}) along an axis decoupled from final task accuracy.
Detailed statistics, sample annotations, and per-VLM breakdowns are deferred to the Appendix.

\section{Conclusion}
\label{sec:conclusion}

We presented \ours{} (\oursshort), a closed-loop framework that pairs a VLM with a video generation model at step-level granularity: the VLM plans one sub-action at a time, inspects the clip the VGM produces, and adaptively chooses among accepting, regenerating, or further decomposing the action.
This step-level coupling redirects test-time compute from sampling more videos toward refining the one being constructed, and it consistently improves both open-source (\vbvrwan) and closed-source (\veo) generators over single-inference, Pass@$k$, and prior test-time scaling baselines on \genvire{} and \vbvr{}.
A human-annotated benchmark further confirms that the VLM's plan-depth, verification, and evolution decisions align with expert annotators, supporting the use of a single VLM as an end-to-end supervisor for the loop.

\paragraph{Limitations.}
Test-time orchestration cannot overcome a VGM that lacks the underlying capability: abstract or symbolic transformations stay hard because the generator never approximates them (Section~\ref{sec:cat_module}), and our gains diminish on lower-capability VGMs whose weak per-step instruction-following and poor preservation of the input frame compound errors across sub-clips faster than re-generation can repair (Section~\ref{sec:appendix_cosmos}).
For the same reason, splitting a task into more steps does not extend the method to longer horizons (Appendix~\ref{sec:appendix_long_horizon}).
Progressive planning on its own also helps only generators that execute the requested intermediate states literally (Section~\ref{sec:ablation}).
The verifier is also imperfect, allowing a fraction of failed clips to propagate downstream (Section~\ref{sec:human_benchmark}).
Future directions include reasoning-oriented VGM training and finer-grained failure localization, which can be orthogonally integrated into our test-time loop.

\clearpage
\begin{ack}
This work was supported by the Institute of Information \& Communications Technology Planning \& Evaluation (IITP) grant funded by the Korea government (MSIT) (No. RS-2024-00457882, AI Research Hub Project; No. RS-2025-02305581, Development of Vision-Language Model (VLM)-Based Intelligent Video Security Monitoring Technology; No. RS-2019-II190075, Artificial Intelligence Graduate School Program (KAIST)).
It was also supported by the National Research Foundation of Korea (NRF) grant (No. RS-2023-00209060, A Study on Optimization and Network Interpretation Method for Large-Scale Machine Learning) funded by the Korea government (MSIT).
\end{ack}

\bibliographystyle{plainnat}
\bibliography{references}

\newpage

\appendix
\begin{LARGE}
    \textbf{Appendix}
\end{LARGE}

\section{Implementation Details}
\label{sec:appendix_implementation}

\subsection{Hyperparameters}
\label{sec:appendix_hyperparams}

\paragraph{Pipeline budget.}
We set $N_{\max}{=}3$ planning steps and $M{=}3$ per-step generation attempts as the default \oursshort{} configuration.
The step-level verifier receives one frame per second uniformly sampled from each candidate clip.
For Gemini 2.5 Pro we instead pass the raw video as input and let the model perform its own frame sampling internally.

\paragraph{Vision-Language Models (VLMs).}
All VLM calls (planner, step-level verifier, and failure router) use Gemini 2.5 Pro (\texttt{gemini-2.5-pro}) with decoding temperature $0.2$.
The same model identity also serves as the output reward model for Pass@$k$ selection and as the critic for the VideoTPO baseline, keeping comparisons matched in VLM compute.
Prompt templates are listed in Appendix~\ref{sec:appendix_prompts}.

\paragraph{Video Generation Models (VGMs).}
For \vbvrwan{}, we use the released $14$B image-to-video checkpoint (Wan2.2 fine-tuned on \vbvr{} reasoning data), generating at a maximum area of $832\times 480$ with aspect ratio preserved, at $16$ fps, with $20$ sampling steps and CFG scale $5.0$.
On \genvire{}, the first step produces a $6$s clip ($96$ frames) and subsequent steps produce $3$s clips ($48$ frames).
On \vbvr{}, we follow the official setup and match each clip to the duration of the corresponding ground-truth video.
Inference runs on a single A100 GPU.
For \veo{}, we use the \texttt{veo-3.1-fast-generate-preview} API at native resolution ($16{:}9$ with letterboxing for non-matching inputs) and $24$ fps.
The first step generates $8$s of video and subsequent steps generate $4$s.
Due to API cost, we evaluate \veo{} only on the smaller \genvire{} benchmark.
For Cosmos-Predict-2.5, we use the released $14$B post-trained checkpoint at $832\times 480$ resolution and $16$ fps, generating $5$s clips per step.
All other settings (sampling steps, guidance scale, negative prompt, scheduler, precision) follow the upstream default configuration of the released checkpoint.

\paragraph{Baselines.}
\emph{Single Inference} generates one video per sample using the same clip length as the $N{=}1$ branch of \oursshort{}.
\emph{Pass@$k$} with $k\in\{2,4\}$ runs $k$ independent generations with seeds $\{1,\ldots,k\}$ under identical inference settings.
Gemini 2.5 Pro is shown all $k$ candidates in a single call and selects the best, acting as an output reward model.
We also tested scoring each seed independently and choosing the highest scorer, but the single-call joint selection gave both higher accuracy and stronger alignment with human judgments.
We expect independent scoring to become more competitive as $k$ grows.
\emph{VideoTPO} follows the official setup with two prompt-rewrite iterations, two seeds per iteration, and a final iteration using only seed $1$, yielding five total generations per sample.
The LLM optimizer and the critic VLM are both Gemini 2.5 Pro to match \oursshort{}.

\paragraph{Evaluation.}
\genvire{} is judged by Gemini 2.5 Pro using the rubric prompts from the official repository, with each sample scored across three independent runs to absorb judge stochasticity.
\vbvr{} uses a deterministic rule-based protocol that compares each generated video against ground-truth references, so we report a single run.
We report Cost as the total VGM-generated seconds per sample, since VLM compute is negligible relative to VGM compute.

\subsection{Prompt Templates}
\label{sec:appendix_prompts}
We provide the verbatim prompt templates used by the VLM in our pipeline:
(i) the \emph{progressive planner}, which emits the next action prompt and a task-complete flag; and
(ii) the \emph{step verifier}, which returns an accept/reject judgment with a structured diagnosis $d$ (textual reason, actionable suggestion, and a \texttt{good\_fraction} estimate of how much of the clip executed correctly).
Each VLM call additionally receives the input image and the history of accepted clip frames; the verifier additionally receives the candidate clip.
Prompt evolution itself does not require a separate VLM call: the verifier's \texttt{suggestion} field is folded directly into the next action prompt by $\mathrm{evolve}(a_t, d)$.

\begin{promptbox}[Progressive planner (\texttt{step\_planner})]
\begin{Verbatim}[fontsize=\small,commandchars=\\\{\},breaklines=true,breakanywhere=true]
You are a visual reasoning expert that plans video generation steps for an image-to-video model.

Given an input image and a task prompt, you must break the task into sequential action steps that a video generation model can execute one at a time. Each step should describe a short, concrete visual action (4-8 seconds of video).

Inputs
- TASK_PROMPT: The full task description
- CURRENT_IMAGE: The current state (input image or last frame of previous clip)
- COMPLETED_STEPS: What has already been done (empty if first step)
- STEP_NUMBER: Which step we are planning (1-indexed)

Rules
1. Each step must describe ONE clear visual action that a video generation model can simulate in 6 seconds.
2. DO NOT plan the entire task at once. Only plan the NEXT IMMEDIATE step.
3. Describe the action in terms of VISIBLE MOTION and CHANGE -- what should move, where, and how.
4. Include the EXACT target state: what the frame should look like when this step is done.
5. If the task appears to be already complete based on the current image, set "task_complete" to true.

Output (strict JSON):
\{
  "observation": "Brief description of what you see in the current image",
  "remaining_goal": "What still needs to happen to complete the task",
  "task_complete": false,
  "instruction": "Detailed video generation prompt for the next step.",
  "target_state": "Visual description of what the last frame should look like after this step",
  "estimated_steps_remaining": 2
\}
\end{Verbatim}
\end{promptbox}

\begin{promptbox}[Step verifier (\texttt{step\_verifier})]
\begin{Verbatim}[fontsize=\small,commandchars=\\\{\},breaklines=true,breakanywhere=true]
You are a video quality judge evaluating whether a generated video clip executed its intended action correctly.

Inputs: TASK_PROMPT, PLANNED_ACTION, TARGET_STATE, VIDEO.

Critical Distinction: judge whether the ACTION was executed, NOT whether the full task is complete.

Evaluation Criteria
1. Did the intended motion/transformation START to happen in the correct direction?
2. Is the result CONSISTENT with the planned action (even if incomplete)?
3. Were there FUNDAMENTAL errors (wrong direction, wrong object, completely wrong action, scene collapse)?

What is NOT a rejection reason:
- Action happened but didn't fully complete (partial progress is fine)
- Minor rendering artifacts or small imprecisions
- The final task goal is not yet reached (planner's job)

On Rejection -- estimate "good_fraction" (0.0-1.0): the fraction of the video that was correct before the error occurs. We use this to do partial re-generation from the good portion.

Output (strict JSON):

If accept:
\{
  "verdict": "accept",
  "action_executed": true,
  "progress": "partial or complete",
  "confidence": "high",
  "reason": "...",
  "suggestion": ""
\}

If reject:
\{
  "verdict": "reject",
  "action_executed": false,
  "progress": "none",
  "good_fraction": 0.3,
  "confidence": "high",
  "reason": "...",
  "suggestion": "..."
\}
\end{Verbatim}
\end{promptbox}

\begin{figure}[ht]
    \centering
    \vspace{-0em}
    \includegraphics[width=\linewidth]{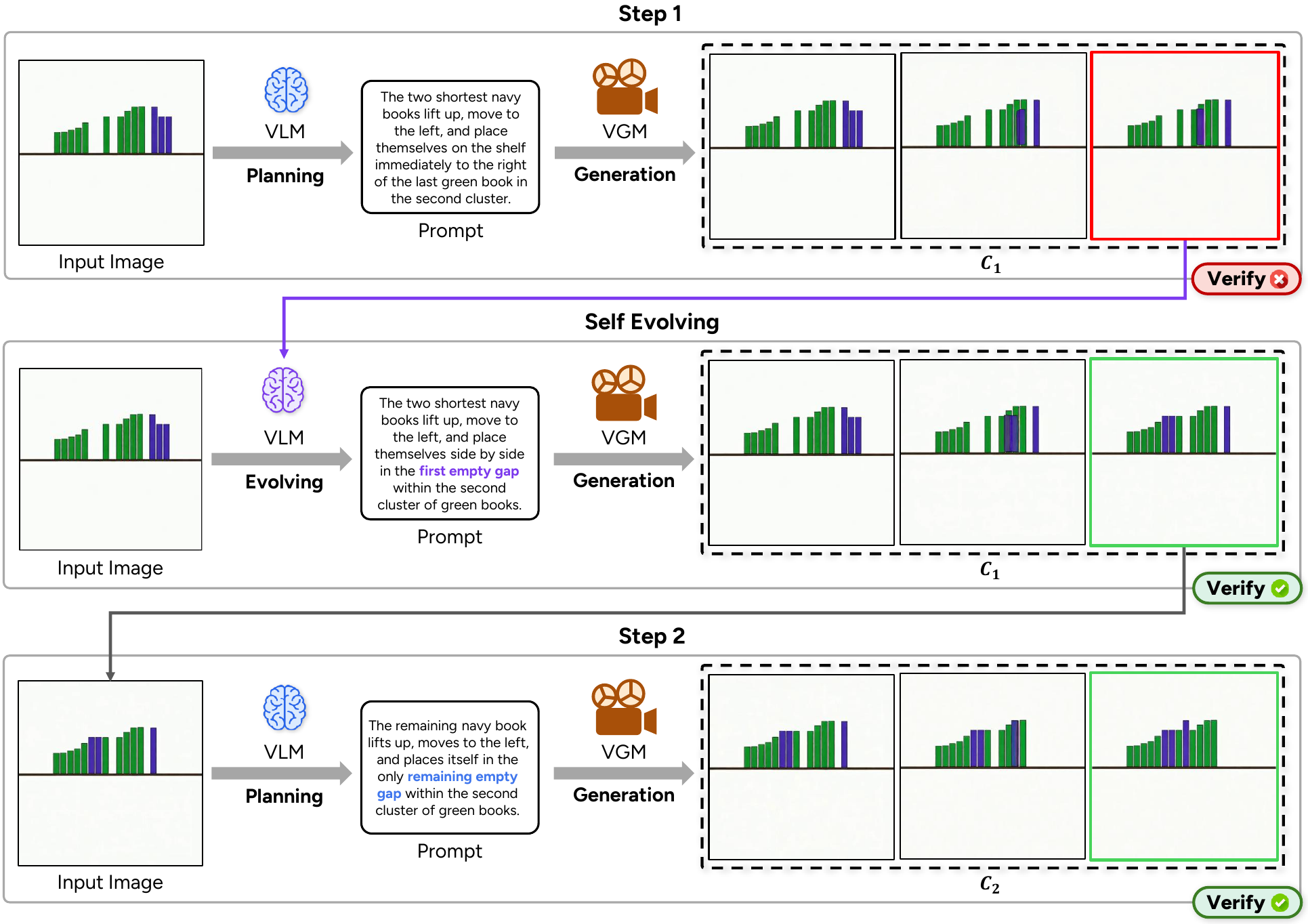}
    \caption{
\textbf{Trace through one full \oursshort{} loop on a multi-step bookshelf task (\vbvrwan{}).}
Within-step prompt evolution (M2) corrects $C_1$ in Step~1, after which across-step progressive planning (M1) advances to Step~2, sharing the same per-clip verifier.
}
    \label{fig:app:VBVR_Evolution}
\end{figure}

\subsection{Verifier Output Examples}
\label{sec:appendix_verifier_examples}
Figure~\ref{fig:app:VBVR_Evolution} instantiates one full execution of Algorithm~\ref{alg:collabvr} on a multi-step bookshelf task, exercising both branches of the inner loop in a single trace.

\paragraph{Step~1: the reject-and-evolve branch.}
The planner emits $a_1$ (\emph{``\dots place the two navy books on the shelf immediately to the right of the last green book\dots''}) and the VGM produces $c_1$.
The verifier rejects: the two navy books land flush against the cluster's right edge rather than inside an empty gap.
Its diagnosis $d_1$, packaging a textual reason and a concrete suggestion, is folded into the next prompt as $a_1' = \mathrm{evolve}(a_1, d_1)$ (\emph{``\dots first empty gap within the second cluster\dots''}, the Self-Evolving panel).
The VGM regenerates $c_1$, and the verifier now accepts; the clip is committed to $\mathcal{H}$ and its last frame becomes the conditioning frame for Step~2.

\paragraph{Step~2: the accept-on-first-attempt branch.}
Conditioned on $\mathcal{H} = \{c_1\}$, the planner emits $a_2$ for the remaining navy book.
The VGM produces $c_2$, the verifier accepts immediately, and the outer loop terminates at $N{=}2$ without ever entering the $\mathrm{evolve}$ branch.

\paragraph{Coverage of the $(v, d)$ interface.}
Together, the three verifier calls cover both routes the per-clip $(v, d)$ interface can take: \texttt{reject}-with-suggestion drives within-step prompt rewriting (M2), and \texttt{accept} commits the clip and advances the across-step plan (M1).
The auxiliary failure router (Appendix~\ref{sec:appendix_failure_router}) is reached only when within-step recovery is exhausted, which does not happen here.

\subsection{First-Frame Fidelity for VGM Selection}
\label{sec:appendix_sora}
First-frame fidelity, which measures how closely the first frame of a generated clip matches the conditioning image, is a prerequisite for the step-by-step clip concatenation in our pipeline (each step's first frame must faithfully resume from the previous step's final frame) as well as for the partial re-generation auxiliary that re-enters the VGM at a mid-clip frame (Appendix~\ref{sec:maze}); we therefore use it as a key selection criterion for candidate VGMs.
We compute the SSIM between the input image and the first frame of the generated video on \genvire{}.

\begin{table}[ht]
\centering
\caption{First-frame fidelity (SSIM) on \genvire{}.}
\label{tab:first_frame_ssim}
\vspace{0.4em}
\begin{tabular}{lc}
\toprule
VGM & SSIM (mean $\pm$ std) \\
\midrule
\vbvrwan{} & $0.970 \pm 0.043$ \\
Cosmos-Predict 2.5 & $0.971 \pm 0.037$ \\
\veo{} & $0.977 \pm 0.035$ \\
\sora{} & $0.818 \pm 0.137$ \\
\bottomrule
\end{tabular}
\end{table}

\vbvrwan{}, \veo{}, and Cosmos-Predict 2.5 all achieve sufficient first-frame fidelity for our pipeline.
\sora{} uses an \texttt{input\_reference} field rather than strict first-frame conditioning, so its first frame drifts from the input.
We therefore exclude \sora{} from main \oursshort{} results since downstream step concatenation cannot preserve information across clips, and \sora{}'s upcoming service deprecation further reduced the value of additional engineering against it.

\subsection{Auxiliary Failure Router}
\label{sec:appendix_failure_router}
Our implementation extends the per-step prompt evolution of Section~\ref{sec:collab_reasoning} with an auxiliary VLM call, the \texttt{failure\_router}, that takes over when evolution alone cannot produce an accepted clip.
After all $M$ within-step evolution attempts return reject, the router examines the failed clip together with the verifier's diagnosis and chooses one of three follow-up actions.
The simplest is \textit{regen}, a single-shot retry that starts from the first failing frame whenever the verifier's \texttt{good\_fraction} is moderate-to-high, so the correctly executed prefix is preserved (Appendix~\ref{sec:maze}).
When the failure is structural rather than a one-off slip, the router triggers \textit{split}, which decomposes the residual task into additional sub-steps and re-enters the progressive planner.
Finally, when prior decomposition has produced step-boundary artifacts on what is fundamentally a single-shot transformation, \textit{fallback} collapses to a single-inference run with $N{=}1$.

\veo{}'s strong single-shot prior often suffices on its own. We therefore invoke the same router at the sample level before any per-step planning, choosing between accepting the single-shot baseline and proceeding with multi-step orchestration. The full prompt template is shown below.

\begin{promptbox}[Failure router (\texttt{failure\_router})]
\begin{Verbatim}[fontsize=\small,commandchars=\\\{\},breaklines=true,breakanywhere=true]
You are a visual reasoning expert deciding how to recover from a FAILED single-shot video generation. Choose among three recovery strategies.

Inputs: TASK_PROMPT, INPUT_IMAGE, FAILED_VIDEO, REJECT_REASON, SUGGESTION, GOOD_FRACTION.

Strategies:
- "regen": single-shot retry. Choose when the failure was an execution slip on a single coherent transformation; good_fraction moderate-to-high; partial regen will likely fix it.
- "split": decompose into multiple steps. Choose when the failure is structural -- the task fundamentally needs intermediate states (multi-action procedures, navigation with turns, assembly, drawing multiple objects), or good_fraction is very low and the failure indicates conflated steps.
- "fallback": collapse to single-inference (N=1). Choose when prior decomposition introduced step-boundary artifacts and the residual task is simple enough to be generated in a single clip.


Output (strict JSON):
\{"action": "regen", "suggestion": "...", "reason": "..."\}
or
\{"action": "split", "estimated_steps": 3, "suggestion": "...", "reason": "..."\}
or
\{"action": "fallback", "suggestion": "...", "reason": "..."\}
\end{Verbatim}
\end{promptbox}

\section{Additional Quantitative Results}
\label{sec:appendix_results}

\subsection{User Study}
\label{sec:appendix_user_study}
\begin{figure}[ht]
\centering
\includegraphics[width=0.85\linewidth]{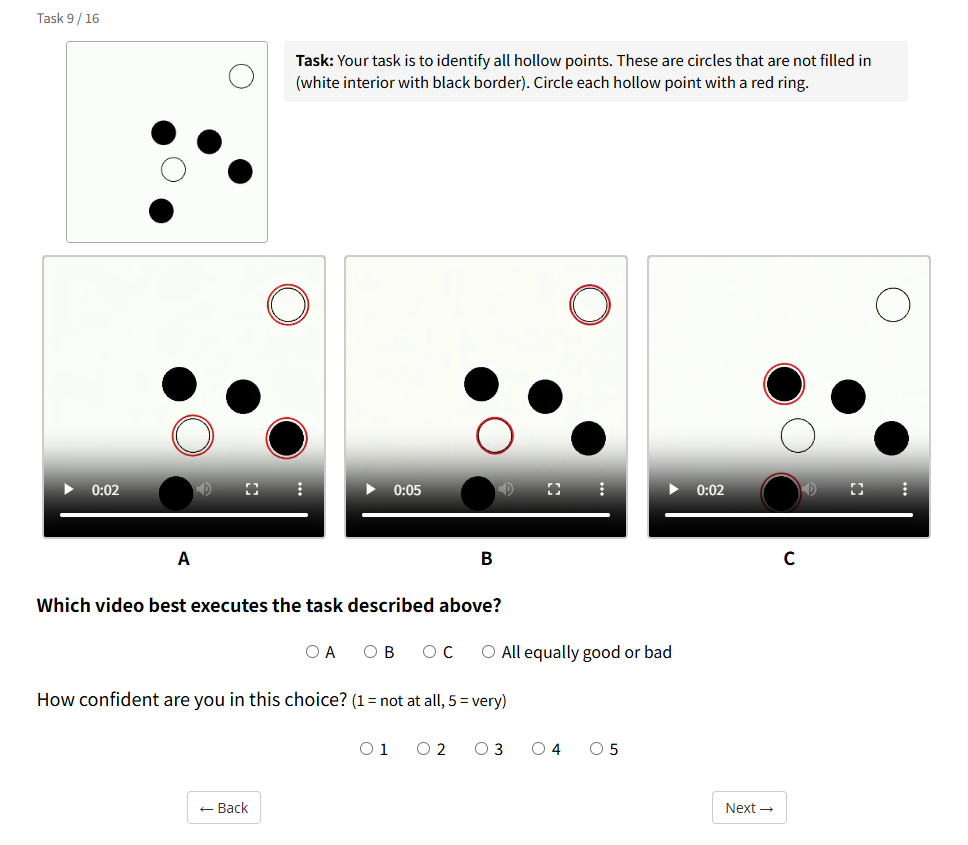}
\caption{\textbf{Per-trial user-study UI.} The participant sees the task prompt and the input image, watches three blinded videos, and answers a forced-choice preference and a confidence rating. The condition$\to$label mapping is randomized per task per participant.}
\label{fig:app:user_study_screenshot}
\end{figure}

We additionally conduct a blind, side-by-side human study comparing \oursshort{} against Pass@$1$ and Pass@$4$ at matched VGM (\vbvrwan{}). The study is hosted on Prolific, and the per-trial UI is shown in Figure~\ref{fig:app:user_study_screenshot}.

\paragraph{Sample selection.}
We curate $16$ tasks spanning \genvire{} and \vbvr{} (one task per Gen-ViRe category and a balanced In-Domain / Out-of-Domain split on VBVR). Within each category we pick the task whose per-sample score gap between \oursshort{} and Pass@$1$/Pass@$4$ is largest (top-$|\Delta|$), excluding cases where all three methods score $0$ or $1$ (no signal). For each task, three videos (Pass@$1$, Pass@$4$, \oursshort{}) are mapped to the labels A/B/C by a fresh random permutation per task per participant.

\paragraph{Procedure.}
On each task page, the participant sees the input image, the task description, and three blinded videos, and answers two questions: ``which video best executes the task?'' (A/B/C/equal) and a $1$--$5$ confidence Likert.

\begin{figure}[ht]
\centering
\includegraphics[width=0.85\linewidth]{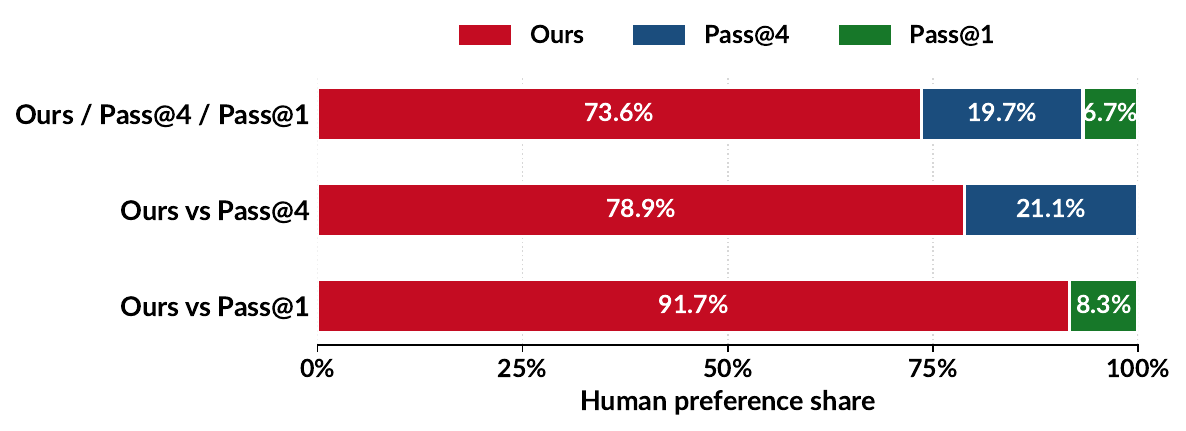}
\caption{\textbf{Human preference share on the user study} ($n{=}40$ participants, $16$ tasks; \emph{Equal} responses excluded). Each row is a $100\%$-stacked bar restricted to the listed conditions. \oursshort{} is the dominant choice in all three views.}
\label{fig:app:user_study_winrate}
\end{figure}

\paragraph{Results.}
We collect $n{=}40$ valid submissions across the $16$ tasks. \emph{Equal} responses comprise $17.7\%$ of all trials; aggregating the remaining decisive preferences:
In head-to-head comparisons (excluding \emph{Equal} votes), \oursshort{} is preferred over Pass@$1$ in $\mathbf{91.7\%}$ and over Pass@$4$ in $\mathbf{78.9\%}$ (Figure~\ref{fig:app:user_study_winrate}). These margins corroborate the automated-metric gains in Section~\ref{sec:experiments}: human raters on a blinded side-by-side comparison consistently prefer \oursshort{} outputs over both single-shot and best-of-$4$ baselines. Inter-rater agreement is moderate: average pairwise raw agreement is $66.3\%$ (decisive-only) and Gwet's AC$_1$ is $0.575$.

\subsection{Effect of Per-Step Attempt Budget \texorpdfstring{$M$}{M}}
\label{sec:appendix_M_ablation}

\begin{wrapfigure}{r}{0.5\linewidth}
    \vspace{-1em}
    \centering
    \includegraphics[width=\linewidth]{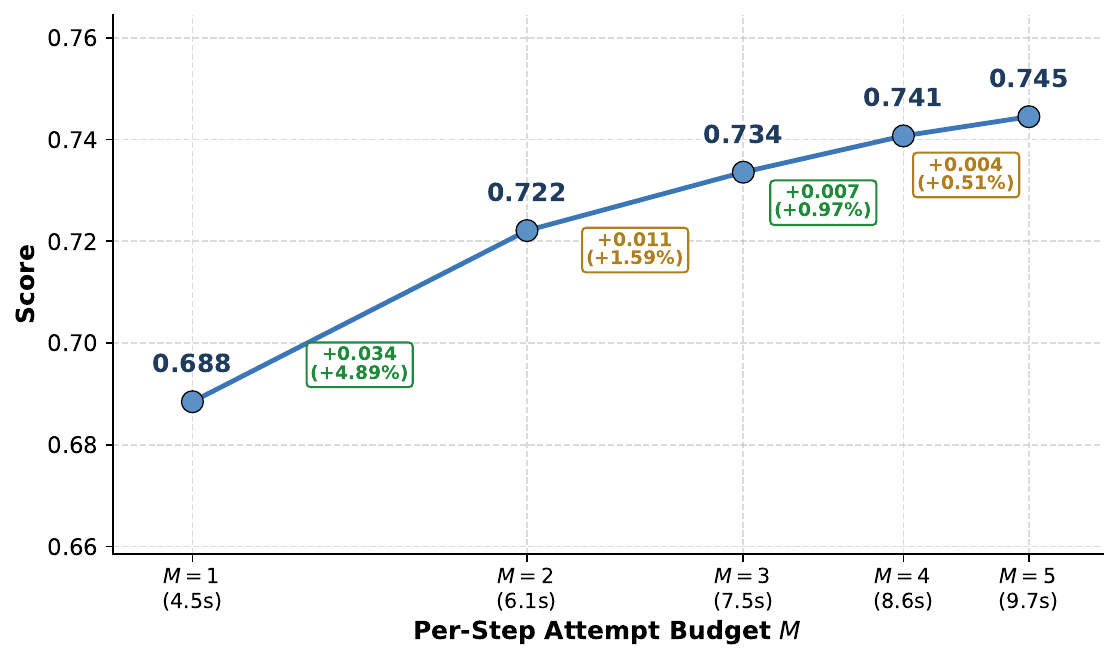}
    \caption{
        \textbf{Per-step attempt budget $M$ on \vbvr{} (\vbvrwan{}, M2-only).}
        Score grows monotonically with $M$, but per-step gains drop below $1\%$ beyond $M{=}3$ while cost continues to scale nearly linearly.
    }
    \vspace{-1em}
    \label{fig:app:VBVR_VBVRWan2.2_by_M}
\end{wrapfigure}

\paragraph{Effect of attempt budget $M$.}
We sweep the per-step attempt budget $M \in \{1, \ldots, 5\}$ on \vbvr{} with \vbvrwan{} under the M2-only configuration (Figure~\ref{fig:app:VBVR_VBVRWan2.2_by_M}).
The first two budget increments deliver most of the gain ($+4.89\%$ from $M{=}1$ to $M{=}2$, $+1.59\%$ from $M{=}2$ to $M{=}3$), after which each additional attempt yields below $1\%$ ($+0.97\%$, $+0.51\%$) while cost continues to grow at a near-constant per-step rate.
Beyond $M{=}3$, the cost-quality profile thus collapses to that of plain Pass@$k$ resampling: extra compute buys little more than an additional independent draw.
We therefore set $M{=}3$ as the default, retaining the cost-efficient regime where verifier-guided re-generation still provides meaningful repair.

\subsection{Per-VLM Human-Alignment Breakdown}
\label{sec:appendix_per_vlm_breakdown}
We provide the per-VLM, per-axis numbers underlying Section~\ref{sec:human_benchmark} in Table~\ref{tab:human_eval_pooled}, pooling annotations from \genvire{} and \vbvr{} (the latter contributed by collaborators).
Three axes are evaluated: D1 plan-depth ($N$ prediction), D2 verifier accept/reject agreement on step clips, and D3 evolution quality of the verifier's suggested repair (1=poor, 2=moderate, 3=well).

\begin{table}[ht]
\centering
\caption{Overall human-eval across 3 VLMs.
D1 plan-depth on VBVR ($n{=}100$, GT video shown) and \genvire{} ($n{=}72$, image only).
D2 verifier agreement on a balanced $125$ accept + $125$ reject sub-sample.
D3 evolution-suggestion quality on $n{=}80$ reject clips per VLM.}
\label{tab:human_eval_pooled}
\vspace{0.4em}
\begin{tabular}{lccc}
\toprule
& Gemini 2.5 Pro & Qwen3.5-27B & Qwen3.5-9B \\
\midrule
\multicolumn{4}{l}{\emph{D1 plan-depth (exact-match)}} \\
\quad \vbvr{} & 64.0\% & 64.2\% & 53.0\% \\
\quad \genvire{} & 73.6\% & 55.6\% & 61.1\% \\
\quad \textbf{Overall} & \textbf{68.0\%} & 58.7\% & 56.4\% \\
\quad MAE (parseable) & 0.366 & 0.491 & 0.484 \\
\midrule
\multicolumn{4}{l}{\emph{D2 verifier agreement}} \\
\quad accept-recall & 84.8\% & 82.4\% & 77.6\% \\
\quad reject-recall (failure detection) & \textbf{65.6\%} & 44.8\% & 40.8\% \\
\quad \textbf{Overall} & \textbf{75.2\%} & 63.6\% & 59.2\% \\
\quad Cohen's $\kappa$ (\genvire) & 0.676 & 0.432 & 0.304 \\
\midrule
\multicolumn{4}{l}{\emph{D3 evolution quality (mean 1--3)}} \\
\quad \textbf{Overall mean} & \textbf{2.61} & 2.55 & 2.35 \\
\quad $\geq 2$ (moderate or better) & 93.8\% & 95.0\% & 86.3\% \\
\quad $=3$ (top) & 67.5\% & 60.0\% & 48.8\% \\
\bottomrule
\end{tabular}
\end{table}

Gemini 2.5 Pro leads on every axis.
The closed-vs-open gap is largest on the verifier (D2 reject-recall: $65.6\%$ vs.\ $40$--$45\%$, a $\sim 21$-point gap; per-verdict confusion matrices in Table~\ref{tab:d2_confusion_matrices}), suggesting the gap is in training data and objective rather than scale alone (the size lift from 9B to 27B contributes only $+4$ points on this axis).
On the evolution axis (D3) the closed-vs-open gap shrinks to $\Delta = 0.06$ (Gemini--27B), within noise.
D1 ranking flips between setups: 9B over-counts steps when given the GT video on \vbvr{} (N=3 collapses to 2/20 correct) but matches Gemini's plan-depth distribution on image-only \genvire{}.

\begin{table}[ht]
\centering
\caption{Verdict confusion matrices, balanced 125:125 sub-sample.}
\label{tab:d2_confusion_matrices}
\vspace{0.4em}
\begin{tabular}{l ccc ccc ccc}
\toprule
& \multicolumn{2}{c}{Gemini 2.5 Pro} && \multicolumn{2}{c}{Qwen3.5-27B} && \multicolumn{2}{c}{Qwen3.5-9B} \\
\cmidrule(lr){2-3} \cmidrule(lr){5-6} \cmidrule(lr){8-9}
& \multicolumn{1}{c}{H=acc} & \multicolumn{1}{c}{H=rej} && \multicolumn{1}{c}{H=acc} & \multicolumn{1}{c}{H=rej} && \multicolumn{1}{c}{H=acc} & \multicolumn{1}{c}{H=rej} \\
\midrule
pred=accept & 106 & 43 && 103 & 69 && 97 & 74 \\
pred=reject &  19 & 82 &&  22 & 56 && 28 & 51 \\
\bottomrule
\end{tabular}
\end{table}

\subsection{Per-Category \texorpdfstring{$\Delta$}{Delta} Heatmap on \texorpdfstring{\vbvr{}}{VBVR-Bench}}
\label{sec:appendix_per_category_heatmaps}

\begin{wrapfigure}{r}{0.5\linewidth}
    \vspace{-1em}
    \centering
    \includegraphics[width=\linewidth]{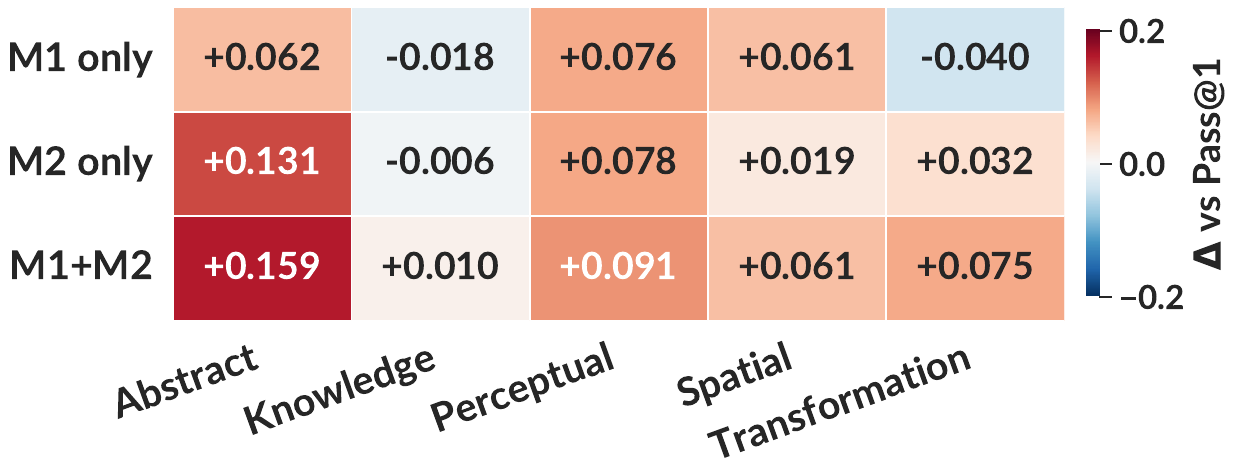}
    \caption{
\textbf{Per-category $\Delta$ over Pass@1 on \vbvr{} (\vbvrwan{}).}
Module configurations follow Section~\ref{sec:ablation}; per-category overall is the sample-count-weighted mean of the In-Domain and Out-of-Domain entries (categories carry $115$, $70$, $150$, $65$, $100$ samples with different ID/OOD splits).
}
    \label{fig:category_module_vbvr}
\end{wrapfigure}

We mirror the per-category $\Delta$ analysis of Section~\ref{sec:cat_module} on \vbvr{} (Figure~\ref{fig:category_module_vbvr}).
Module dominance flips relative to \genvire{}: M2 alone is now the larger single-module contributor (Table~\ref{tab:method_ablation}: $\Delta_{\text{M2}}\,{+}0.063$ vs.\ $\Delta_{\text{M1}}\,{+}0.035$), driven primarily by Abstract (M2 alone $+0.131$) where atomic perceptual targets are recoverable through verifier-driven re-sampling but admit no genuine sub-goals to expose.
The two modules dominate different categories: M1 alone is actually the larger contributor on Spatial ($+0.061$ vs.\ M2~$+0.019$), where decomposition exposes navigation-style sub-goals that the VGM can satisfy individually, while Transformation is the clearest case in which neither module alone is sufficient ($\Delta_{\text{M1}}\,{-}0.040$, $\Delta_{\text{M2}}\,{+}0.032$) yet the combination recovers a clean positive gain ($\Delta_{\text{M1+M2}}\,{+}0.075$).
With both modules enabled, all five categories turn positive ($+0.010$ to $+0.159$), reproducing the per-instance adaptive activation observed on \genvire{}.

Yet the symbolic limit identified in Section~\ref{sec:cat_module} surfaces here as well: Knowledge remains marginal across all three configurations ($\Delta\,{+}0.010$ for M1+M2, an order of magnitude below the other categories), since neither decomposing nor re-sampling supplies the missing world knowledge---there is no sub-step to expose, and the verifier can only redraw from the VGM's existing distribution.
This parallels the small Analogy ($+0.083$) and Abstract ($+0.090$) gains on \genvire{}, again pointing to knowledge-grounded VGM training as a complementary direction rather than additional planner or verifier modules.

\subsection{Veo 3.1 Module Ablation}
\label{sec:appendix_veo_ablation}

We ablate the same M1 / M2 modules on \genvire{} with \veo{}, mirroring Table~\ref{tab:method_ablation} for \vbvrwan{}.

\begin{table}[ht]
\centering
\small
\caption{Per-module ablation on \genvire{} with \veo{}. $\Delta$ over Pass@1. M1 / M2 columns indicate which module is enabled. The last row is an additional ablation that fixes $N{=}3$ for every sample, bypassing the planner's per-state adaptive $N$ selection.}
\label{tab:veo_module_ablation}
\vspace{0.4em}
\setlength{\tabcolsep}{7pt}
\begin{tabular}{ccccc}
\Xhline{3\arrayrulewidth}
\rule{-2.5pt}{10.0pt}
\bf M1 & \bf M2 & \bf Cost (s) & \bf Overall & \bf $\Delta$ \\
\midrule
\textcolor[HTML]{A30000}{\xmark} & \textcolor[HTML]{A30000}{\xmark} & 8.0  & 0.481 & -- \\
\textcolor[HTML]{007304}{\cmark} & \textcolor[HTML]{A30000}{\xmark} & 10.1 & 0.446 & $-$0.035 \\
\textcolor[HTML]{A30000}{\xmark} & \textcolor[HTML]{007304}{\cmark} & 12.5 & 0.527 & +0.046 \\
\textcolor[HTML]{007304}{\cmark} & \textcolor[HTML]{007304}{\cmark} & 21.4 & \bf 0.550 & \bf +0.069 \\
\midrule
\multicolumn{5}{l}{\emph{Ablation: fixed $N{=}3$ (bypass progressive $N$ selection)}} \\
\textcolor[HTML]{007304}{\cmark} & \textcolor[HTML]{007304}{\cmark} & 24.5 & 0.450 & $-$0.031 \\
\Xhline{3\arrayrulewidth}
\end{tabular}
\end{table}

\paragraph{The verifier-and-regen module M2 alone has a single-shot ceiling.}
With $N{=}1$ and $M{=}2$ (no decomposition, regen only), \veo{} reaches $0.527$, a $+0.046$ gain over Pass@1 that is almost identical to the corresponding $+0.045$ gain on \vbvrwan{} (Table~\ref{tab:method_ablation}).
However, this number reflects only the tasks that \veo{} can complete in a single clip; tasks requiring multiple sequential sub-actions remain out of reach for any number of regenerations.

\paragraph{M1 unlocks the tasks where single-shot is insufficient.}
The full M1+M2 configuration (max $N{=}3$, max $M{=}2$) reaches $0.550$, a further $+0.023$ over the M2-only ceiling.
The gain comes from samples on which the single-shot stream cannot complete the task: there, the planner emits a multi-step decomposition ($N{>}1$) that breaks the task into sub-actions \veo{} can satisfy individually.
M1's decomposition is therefore the active ingredient: it unlocks tasks that no amount of regeneration on a single clip could solve.
Even with the maximum decomposition forced ($N{=}3$ for every sample, last row of Table~\ref{tab:veo_module_ablation}), \veo{} drops to $0.450$, confirming that adaptive $N$ selection, not just the existence of decomposition, is what makes M1 productive.

\paragraph{Why M1 alone helps \vbvrwan{} but actually \emph{hurts} \veo{}.}
M1 alone delivers most of the total gain on \vbvrwan{} ($+0.120$ of $+0.140$; Table~\ref{tab:method_ablation}), yet on \veo{} it falls $0.035$ below Pass@1.
The asymmetry follows from the two VGMs' design priors.
\vbvrwan{} is fine-tuned for visual reasoning, so the artificial intermediate states our planner emits (partial rotations, half-drawn lines, incremental fills) are inside its training distribution; each sub-step executes literally, leaving the next step's conditioning aligned with the planner's target.
\veo{}'s strong end-to-end motion prior gives it a higher single-shot Pass@1 ($0.481$ vs.\ $0.391$ for \vbvrwan{}), but reinterprets these intermediate-state requests into outputs the prior considers natural, so with $M{=}0$ the deviation feeds the next step's conditioning unchecked and errors compound across the chain.
M1 alone therefore hurts \veo{} most on Planning ($-0.069$) and Spatial ($-0.067$), the categories where \veo{}'s single-shot capacity was highest to begin with.
M2 reconciles \veo{}'s generalist prior with the planner's task-specific structure: the verifier rejects mid-state deviations before they propagate, and the regen loop pulls \veo{}'s output back onto the trajectory.
This is why M1+M2 reaches $0.550$ even though M1 alone underperforms, and the same dynamic explains the Spatial-category synergy in Section~\ref{sec:cat_module}: long-horizon simulation needs both modules, since M1 alone accumulates deviation while M2 alone hits the single-shot ceiling.

\paragraph{Leave-one-out ablation.}
Table~\ref{tab:leave_one_out} reports the complementary ablation, in which one module is removed from the full pipeline.
Removing either module lowers the score in all three settings.
On \veo{}, removing M1 costs $0.023$ even though M1 alone is below Pass@1, because M1 lets the full pipeline solve tasks that cannot be completed within a single clip.

\begin{table}[ht]
\centering
\small
\caption{Leave-one-out ablation. Each entry is the score after removing one module from the full pipeline, with the difference in parentheses.}
\label{tab:leave_one_out}
\vspace{0.4em}
\setlength{\tabcolsep}{7pt}
\begin{tabular}{lccc}
\Xhline{3\arrayrulewidth}
\rule{-2.5pt}{10.0pt}
\bf Setting & \bf Full & \bf w/o M1 & \bf w/o M2 \\
\midrule
\genvire{}, \vbvrwan{} & 0.531 & 0.436 ($-$0.095) & 0.511 ($-$0.020) \\
\vbvr{}, \vbvrwan{}    & 0.757 & 0.734 ($-$0.023) & 0.706 ($-$0.051) \\
\genvire{}, \veo{}     & 0.550 & 0.527 ($-$0.023) & 0.446 ($-$0.104) \\
\Xhline{3\arrayrulewidth}
\end{tabular}
\end{table}

\subsection{Cosmos-Predict-2.5 Detailed Results}
\label{sec:appendix_cosmos}
\oursshort{} improves Cosmos-Predict 2.5 (14B) on \vbvr{} (Table~\ref{tab:vbvr_results}, $0.308 \to 0.403$) but degrades on \genvire{} ($0.287 \to 0.182$, $\Delta=-0.105$ vs.\ Pass@4; Table~\ref{tab:cosmos_per_cat}).
This contrast follows each benchmark's task profile: \vbvr{} is dominated by single-step tasks, where \oursshort{} mostly invokes single-clip verification with at most a few re-generations, both of which Cosmos handles competently.
\genvire{} is dominated by multi-step reasoning tasks (Section~\ref{sec:ablation}), so \oursshort{} forces Cosmos to execute decomposed sub-actions across $N{\le}3$ clips.
Here Cosmos's weaker per-step instruction-following becomes the bottleneck and decomposition compounds rather than absorbs the error across clips.

\begin{table}[ht]
\centering
\caption{Cosmos-Predict 2.5 per-category on \genvire{}, fair intersection of \oursshort{} and Pass@$k$.}
\label{tab:cosmos_per_cat}
\vspace{0.4em}
\resizebox{\linewidth}{!}{%
\begin{tabular}{l cccccc c}
\toprule
Method & Abstract & Algo & Analogy & Perc. & Planning & Spatial & Avg \\
\midrule
Pass@1 (baseline) & 0.204 & 0.346 & 0.042 & 0.305 & 0.352 & 0.200 & 0.246 \\
Pass@4 & 0.278 & 0.432 & 0.042 & 0.288 & 0.444 & 0.217 & 0.287 \\
\oursshort{} ($N{=}3$) & 0.009 & 0.249 & 0.375 & 0.000 & 0.368 & 0.131 & 0.182 \\
\bottomrule
\end{tabular}%
}
\end{table}

The single \genvire{} category where \oursshort{} still helps Cosmos is Analogy ($0.042 \to 0.375$, $\Delta=+0.333$): Pass@1 is essentially zero there, so any successful sub-step accumulates to a positive end-state.
This delineates the regime where decomposition still helps a weak VGM: when single-shot performance is so low that any partial progress beats the baseline.

\paragraph{Scene collapse in Cosmos-Predict-2.5.}
To find where Cosmos breaks down, we manually inspected all of its clips on \genvire{}, including re-generations.
Each clip was shown as four frames and mixed with \vbvrwan{} clips so that the annotator did not know which model had produced it, and was labeled as collapsed, severely damaged, or intact.
The most frequent failure is scene collapse, in which the clip loses the input frame and turns into an unrelated scene.
Of the Cosmos clips, $31.7\%$ are collapsed and $58.8\%$ are collapsed or severely damaged, compared with about $1\%$ for \vbvrwan{}.
Because each step starts from the last frame of the previous one, a collapsed clip also corrupts the input of the next step.
This explains why Cosmos degrades on \genvire{}, where most tasks require several steps, while it still improves on \vbvr{} ($+0.095$).
The labels were assigned by a single annotator from still frames.

\paragraph{Limitation.}
\oursshort{}'s gain on a given VGM is bounded below by the VGM's per-step instruction-following reliability.
We frame \oursshort{} as orthogonal to, rather than a substitute for, training stronger VGMs: the framework requires a generator that meets a minimum per-step instruction-following floor before decomposition becomes profitable.

\section{Robustness Analysis}
\label{sec:appendix_robustness}

\gemini{} serves both as the VLM inside \oursshort{} and as the rubric judge of \genvire{}.
We therefore repeat the \genvire{} comparison with a different judge, with a different in-loop VLM, and over multiple generation runs, and we also evaluate larger step budgets on long-horizon tasks.
All experiments in this section use \vbvrwan{}.

\subsection{Evaluation with GPT-5 as the Judge}
\label{sec:appendix_gpt_judge}

We re-score the videos from the main experiment with GPT-5 as the rubric judge, keeping the outputs selected for Pass@$k$ unchanged (Table~\ref{tab:gpt_judge}).
Since the GPT-5 API does not take video input, each video is given as eight uniformly sampled frames, and each sample is scored once.
Scores are lower overall under GPT-5, but the five methods keep the same order as under \gemini{}, and the margin of \oursshort{} over VideoTPO is almost unchanged ($+0.042$ against $+0.043$).

\begin{table}[ht]
\centering
\small
\caption{Results on \genvire{} judged by GPT-5. The videos are the same as in Table~\ref{tab:genvire_results}.}
\label{tab:gpt_judge}
\vspace{0.4em}
\begin{tabular}{lccccccc}
\toprule
Method & Avg. & Abst. & Algo. & Analog. & Perc. & Plan. & Spat. \\
\midrule
Pass@1   & 0.345 & 0.375 & 0.441 & 0.250 & 0.280 & 0.371 & 0.353 \\
Pass@2   & 0.382 & 0.479 & 0.457 & 0.250 & 0.298 & 0.442 & 0.367 \\
Pass@4   & 0.403 & 0.542 & 0.415 & 0.333 & 0.280 & 0.479 & 0.369 \\
VideoTPO & 0.435 & 0.458 & 0.431 & 0.500 & 0.306 & 0.562 & 0.353 \\
\oursshort{} & \textbf{0.477} & 0.406 & 0.620 & 0.417 & 0.323 & 0.654 & 0.444 \\
\bottomrule
\end{tabular}
\end{table}

\subsection{Effect of the In-Loop VLM}
\label{sec:appendix_inloop_vlm}

We next replace the VLM inside every method with GPT-5 or Qwen3.5-27B.
This VLM selects the output for Pass@$k$, critiques the video in VideoTPO, and serves as the planner and verifier in \oursshort{}, while the judge remains \gemini{} (Table~\ref{tab:inloop_vlm}).
\oursshort{} achieves the highest score with all three VLMs.
With GPT-5 in the loop it reaches $0.541$, slightly above the $0.531$ obtained with \gemini{}, so the \gemini{} judge does not favor outputs that were produced with \gemini{} in the loop.

\begin{table}[ht]
\centering
\small
\caption{Results on \genvire{} with different VLMs inside each method, judged by \gemini{}.}
\label{tab:inloop_vlm}
\vspace{0.4em}
\begin{tabular}{llccccccc}
\toprule
In-loop VLM & Method & Avg. & Abst. & Algo. & Analog. & Perc. & Plan. & Spat. \\
\midrule
-- & Pass@1 & 0.391 & 0.479 & 0.415 & 0.250 & 0.261 & 0.554 & 0.387 \\
\midrule
\multirow{4}{*}{\gemini{}} & Pass@2 & 0.398 & 0.576 & 0.437 & 0.278 & 0.257 & 0.481 & 0.357 \\
 & Pass@4 & 0.438 & 0.622 & 0.418 & 0.250 & 0.275 & 0.604 & 0.462 \\
 & VideoTPO & 0.488 & 0.535 & 0.443 & 0.417 & 0.313 & 0.671 & 0.552 \\
 & \oursshort{} & \textbf{0.531} & 0.569 & 0.606 & 0.333 & 0.367 & 0.821 & 0.488 \\
\midrule
\multirow{4}{*}{GPT-5} & Pass@2 & 0.406 & 0.556 & 0.447 & 0.222 & 0.274 & 0.572 & 0.368 \\
 & Pass@4 & 0.423 & 0.656 & 0.387 & 0.250 & 0.272 & 0.564 & 0.409 \\
 & VideoTPO & 0.452 & 0.392 & 0.497 & 0.417 & 0.338 & 0.661 & 0.409 \\
 & \oursshort{} & \textbf{0.541} & 0.507 & 0.608 & 0.472 & 0.502 & 0.682 & 0.491 \\
\midrule
\multirow{4}{*}{Qwen3.5-27B} & Pass@2 & 0.398 & 0.503 & 0.379 & 0.250 & 0.280 & 0.547 & 0.427 \\
 & Pass@4 & 0.403 & 0.611 & 0.345 & 0.250 & 0.260 & 0.564 & 0.388 \\
 & VideoTPO & 0.494 & 0.608 & 0.528 & 0.333 & 0.428 & 0.583 & 0.486 \\
 & \oursshort{} & \textbf{0.510} & 0.562 & 0.540 & 0.417 & 0.267 & 0.773 & 0.503 \\
\bottomrule
\end{tabular}
\end{table}

\subsection{Variance over Generation Runs}
\label{sec:appendix_seeds}

All \genvire{} scores in the paper are averaged over three judge runs, which accounts for variance in judging but not in generation.
We therefore run \oursshort{} with three different generation seeds (Table~\ref{tab:seeds}).
The mean score is $0.538$ with a standard deviation of $0.007$, and every run stays above VideoTPO ($0.488$).

\begin{table}[ht]
\centering
\small
\caption{\oursshort{} on \genvire{} over three generation runs. The last row reports the mean and standard deviation.}
\label{tab:seeds}
\vspace{0.4em}
\footnotesize
\setlength{\tabcolsep}{2.5pt}
\resizebox{\linewidth}{!}{%
\begin{tabular}{lccccccc}
\toprule
Method & Avg. & Abst. & Algo. & Analog. & Perc. & Plan. & Spat. \\
\midrule
Pass@1   & 0.391 & 0.479 & 0.415 & 0.250 & 0.261 & 0.554 & 0.387 \\
Pass@4   & 0.438 & 0.622 & 0.418 & 0.250 & 0.275 & 0.604 & 0.462 \\
VideoTPO & 0.488 & 0.535 & 0.443 & 0.417 & 0.313 & 0.671 & 0.552 \\
\midrule
\oursshort{} (seed 1) & 0.531 & 0.569 & 0.606 & 0.333 & 0.367 & 0.821 & 0.488 \\
\oursshort{} (seed 2) & 0.544 & 0.663 & 0.556 & 0.444 & 0.301 & 0.742 & 0.561 \\
\oursshort{} (seed 3) & 0.537 & 0.649 & 0.556 & 0.444 & 0.297 & 0.726 & 0.550 \\
\oursshort{} (mean) & $0.538$\,{\scriptsize$\pm0.007$} & $0.627$\,{\scriptsize$\pm0.051$} & $0.573$\,{\scriptsize$\pm0.029$} & $0.407$\,{\scriptsize$\pm0.064$} & $0.322$\,{\scriptsize$\pm0.039$} & $0.763$\,{\scriptsize$\pm0.051$} & $0.533$\,{\scriptsize$\pm0.039$} \\
\bottomrule
\end{tabular}%
}
\end{table}

\subsection{Effect of Larger Step Budgets}
\label{sec:appendix_long_horizon}

To examine tasks that need longer decompositions, we take the 52 of the 72 \genvire{} samples that use all three steps under the default $N_{\max}{=}3$ and rerun them with $N_{\max}{=}4$ and $N_{\max}{=}5$ (Table~\ref{tab:nmax_subset}).
Neither setting improves over $N_{\max}{=}3$.
Rejections concentrate on the last step regardless of the budget, with final-reject rates at the last step of $46.2\%$, $51.0\%$, and $61.9\%$ for $N_{\max}{=}3$, $4$, and $5$, and $58\%$ of the rejected cases still fail after all five steps.
These failures arise in executing individual sub-actions, and additional steps do not resolve them.
Decomposition reduces the difficulty of chaining several actions in order, but it does not help when a single sub-action is already beyond what the generator can render, and with current VGMs this trade-off favors about three steps.

\begin{table}[ht]
\centering
\small
\caption{\oursshort{} with larger step budgets on the 52 \genvire{} samples that use all three steps at $N_{\max}{=}3$.}
\label{tab:nmax_subset}
\vspace{0.4em}
\begin{tabular}{lccc}
\toprule
$N_{\max}$ & 3 & 4 & 5 \\
\midrule
\genvire{} subset ($n{=}52$) & \textbf{0.517} & 0.463 & 0.482 \\
\bottomrule
\end{tabular}
\end{table}

\section{Module Diagnostics}
\label{sec:appendix_diagnostics}

\subsection{Pipeline Statistics}
\label{sec:appendix_pipeline_stats}
We aggregate the runtime behaviour of \oursshort{}+\vbvrwan{} on \genvire{} and \vbvr{} under the default configuration ($N_{\max}{=}3$, $M{=}3$, i.e.\ up to two re-generations per step).

\paragraph{Per-sample summary.}
Each \genvire{} sample takes on average $2.56$ planning steps, $1.46$ re-generations, $4.01$ generated clips, and $6.79$ VLM calls; on \vbvr{} the same quantities drop to $1.48$, $0.90$, $2.38$, and $3.85$ respectively (Table~\ref{tab:pipe_summary_gvr}), tracking the shorter trajectories on \vbvr{}.

\begin{table}[!ht]
\centering
\small
\caption{Per-sample runtime summary on \genvire{} ($72$ samples) and \vbvr{} ($500$ samples), under default config.}
\label{tab:pipe_summary_gvr}
\vspace{0.4em}
\begin{tabular}{ll cccc}
\toprule
Quantity & Benchmark & Mean & Std & Median & Range \\
\midrule
\multirow{2}{*}{Steps taken}    & \genvire{} & 2.56 & 0.76 & 3 & $[1, 3]$ \\
                                & \vbvr{}    & 1.48 & 0.77 & 1 & $[1, 3]$ \\
\midrule
\multirow{2}{*}{Re-generations} & \genvire{} & 1.46 & 1.72 & 1 & $[0, 6]$ \\
                                & \vbvr{}    & 0.90 & 1.26 & 0 & $[0, 6]$ \\
\midrule
\multirow{2}{*}{Generated clips}& \genvire{} & 4.01 & 2.18 & 3 & $[1, 9]$ \\
                                & \vbvr{}    & 2.38 & 1.96 & 1 & $[1, 9]$ \\
\midrule
\multirow{2}{*}{VLM calls}      & \genvire{} & 6.79 & 2.49 & 6 & $[2, 12]$ \\
                                & \vbvr{}    & 3.85 & 2.69 & 2 & $[2, 12]$ \\
\bottomrule
\end{tabular}
\end{table}

\paragraph{Step trajectory length.}
The planner terminates early at $N{=}1$ for $16.7\%$ of \genvire{} samples (single-shot accept via task-complete signal), at $N{=}2$ for $11.1\%$, and reaches the cap $N{=}3$ for the remaining $72.2\%$.
On \vbvr{} the trajectory is much shorter (mean $N{=}1.48$ vs.\ $2.56$ on \genvire{}; Table~\ref{tab:pipe_n_dist}), consistent with \vbvr{}'s reasoning-heavy but visually constrained tasks where many problems admit a single closed-form action.

\begin{table}[!ht]
\centering
\small
\caption{Step trajectory length distribution on \genvire{} and \vbvr{}.}
\label{tab:pipe_n_dist}
\vspace{0.4em}
\begin{tabular}{lcccc}
\toprule
$N$ (steps used) & 1 & 2 & 3 (cap) & Mean \\
\midrule
\genvire{} share & 16.7\% & 11.1\% & 72.2\% & 2.56 \\
\vbvr{} share    & 69.8\% & 12.8\% & 17.4\% & 1.48 \\
\bottomrule
\end{tabular}
\end{table}

\paragraph{Re-generation distribution.}
Table~\ref{tab:pipe_regen} reports the per-sample re-generation count distribution on both benchmarks. On \genvire{}, the distribution decays gradually from $0$ to the cap of $6$ regens (mean $1.46$). On \vbvr{}, the distribution is bimodal: a primary mode at $0$ regens ($58.6\%$) from samples the planner finishes in a single shot, and a secondary mode at $2$ regens ($26.4\%$) from multi-step trajectories that fail mid-way (mean $0.90$, contributing $451$ extra clips overall).

\begin{table}[!ht]
\centering
\small
\caption{Re-generation distribution per sample on \genvire{} and \vbvr{}. Each row reports the share of samples whose total re-generation count across the trajectory falls in the given bin.}
\label{tab:pipe_regen}
\vspace{0.4em}
\begin{tabular}{lccccccc|c}
\toprule
Re-gens / sample & 0 & 1 & 2 & 3 & 4 & 5 & 6 & Mean \\
\midrule
\genvire{} ($n{=}72$)   & 48.6\% & 9.7\% & 12.5\% & 13.9\% & 9.7\% & 2.8\% & 2.8\% & 1.46 \\
\vbvr{} ($n{=}500$)      & 58.6\% & 7.6\% & 26.4\% & 2.0\% & 4.2\% & 0.2\% & 1.0\% & 0.90 \\
\bottomrule
\end{tabular}
\end{table}

\paragraph{Per-category breakdown on \genvire{}.}
Algorithmic samples consume the most compute per task ($5.0$ clips on average) due to a higher re-generation rate ($2.50$/sample), while Analogy and Planning converge fastest (Table~\ref{tab:pipe_per_category}).

\begin{table}[!ht]
\centering
\small
\caption{Per-category compute breakdown on \genvire{}.}
\label{tab:pipe_per_category}
\vspace{0.4em}
\begin{tabular}{lcccc}
\toprule
Category & Avg $N$ & Avg re-gens & Avg clips & Avg VLM calls \\
\midrule
Abstract        & 2.25 & 1.75 & 4.00 & 6.67 \\
Algorithmic     & 2.50 & 2.50 & 5.00 & 7.83 \\
Analogy         & 2.00 & 1.08 & 3.08 & 5.67 \\
Perceptual      & 3.00 & 1.50 & 4.50 & 7.50 \\
Planning        & 2.75 & 0.67 & 3.42 & 6.17 \\
Spatial         & 2.83 & 1.25 & 4.08 & 6.92 \\
\bottomrule
\end{tabular}
\end{table}

\subsection{Verifier Run-time Behavior}
\label{sec:appendix_verifier_diag}
We characterize the verifier on the same \genvire{} run.

\paragraph{Step-level outcomes.}
The verifier accepts \textbf{49.5\%} of clips on the first attempt and recovers another \textbf{14.7\%} through re-generation, for a \textbf{64.1\%} final acceptance rate at an average retry depth of \textbf{0.57} per step.
The remaining \textbf{35.9\%} are rejected even after the maximum two re-generations and propagated downstream as the best-of-attempts clip.

\paragraph{Per-step rejection trend.}
Final-reject rate is comparable at the first two steps but rises sharply at the deepest step (Table~\ref{tab:verifier_per_step_reject}).
The step-3 spike reflects cumulative visual drift: each step's first frame is the previous step's last frame, so any mild deformation accepted earlier makes the next step harder to satisfy.
The same pattern shows up in the diminishing returns of $N$ beyond $3$ on \genvire{} (Section~\ref{sec:ablation}) and even more sharply on Cosmos-Predict 2.5 (Appendix~\ref{sec:appendix_cosmos}).
Overall, the verifier is exercised aggressively (about $37\%$ of steps trigger at least one re-generation) without saturating the budget, supporting our default choice of $M{=}3$.

\begin{table}[!ht]
\centering
\small
\caption{Verifier final-reject rate by step index on \genvire{}.}
\label{tab:verifier_per_step_reject}
\vspace{0.4em}
\begin{tabular}{lccc}
\toprule
Step index & Step 1 & Step 2 & Step 3 \\
\midrule
Final reject rate & 31.9\% & 31.7\% & 46.2\% \\
\bottomrule
\end{tabular}
\end{table}

\subsection{Cost Decomposition}
\label{sec:appendix_cost}

We verify the ``VLM compute is negligible relative to VGM compute'' claim in Section~\ref{sec:implementation} along two axes: GPU wall-clock time on the open-source backbone (\vbvrwan{}, single A100) and API \$ cost on the closed-source backbone (\veo{}).

\paragraph{VLM per-call profile.}
We replayed representative \gemini{} calls on \genvire{} (planner: $1$ image $+$ task description; verifier: $1$ instruction $+$ $1$ video clip) and recorded latency and token counts, averaged over $3$ trials per call type.

\begin{table}[ht]
\centering
\small
\begin{tabular}{lccc}
\toprule
Call type & Latency & Input tok & Output tok \\
\midrule
Planner       & $10.15$\,s & $800$ & $186$ \\
Step verifier & $10.60$\,s & $2{,}724$ & $117$ \\
\bottomrule
\end{tabular}
\vspace{0.4em}
\caption{Per-call \gemini{} profile. The verifier's $2.7$\,K-token input is dominated by the inline video; the planner's $800$-token input is the prompt template, the task description, and a single image.}
\label{tab:per_call_vlm}
\end{table}

\begin{table}[ht]
\centering
\small
\begin{tabular}{lcc}
\toprule
Component & Per-sample wall-clock & Share \\
\midrule
\vbvrwan{} VGM (A100) & $979.6$\,s ($16.3$\,min) & $93.5\%$ \\
\gemini{} VLM (planner $+$ verifier) & $68.5$\,s ($1.14$\,min) & $6.5\%$ \\
\midrule
Total & $1048.1$\,s ($17.5$\,min) & $100\%$ \\
VGM / VLM ratio & \multicolumn{2}{c}{$\approx 14\times$} \\
\bottomrule
\end{tabular}
\vspace{0.4em}
\caption{Per-sample wall-clock decomposition (\vbvrwan{}\,+\,\oursshort{} on a single A100, averaged over \genvire{}).}
\label{tab:cost_time}
\end{table}

\paragraph{Closed-source: API cost.}
\veo{} Fast is priced at $\$0.15$/s of generated video, so the per-sample $21.4$\,s reported in Table~\ref{tab:genvire_results} costs $\$3.21$, compared to $\$0.026$ for the VLM, a $\sim 125\times$ ratio (Table~\ref{tab:cost_money}).

\begin{table}[ht]
\centering
\small
\begin{tabular}{lcc}
\toprule
Component & Per-sample \$ & Share \\
\midrule
\veo{} VGM & $\$3.210$ & $99.2\%$ \\
\gemini{} VLM (planner $+$ verifier) & $\$0.026$ & $0.8\%$ \\
\midrule
Total & $\$3.236$ & $100\%$ \\
VGM / VLM ratio & \multicolumn{2}{c}{$\approx 125\times$} \\
\bottomrule
\end{tabular}
\vspace{0.4em}
\caption{Per-sample API cost decomposition (\veo{}\,+\,\oursshort{} on \genvire{}).}
\label{tab:cost_money}
\end{table}

\paragraph{Per-sample VLM aggregate.}
With $2.56$ planner and $4.01$ verifier calls per sample (Section~\ref{sec:appendix_pipeline_stats}), each \genvire{} sample triggers $\sim 6.6$ VLM calls totalling $\sim 68.5$\,s of API wall-clock and $\sim 13$\,K input tokens.
At \gemini{} pricing ($\$1.25$/M input, $\$10$/M output), this works out to $\$0.026$ per sample.

\paragraph{Open-source: GPU wall-clock time.}
On a single A100 at $480$p with $20$ inference steps, \vbvrwan{} generates the per-sample $17.8$\,s of video in approximately $980$\,s of wall-clock, while the VLM contributes only $\sim 68.5$\,s, a $\sim 14\times$ ratio (Table~\ref{tab:cost_time}).

On both axes the VGM dominates by an order of magnitude or more, justifying the convention in Section~\ref{sec:implementation} of reporting Cost as VGM-generated seconds per sample as a faithful proxy for total compute on either deployment regime.

\clearpage

\section{Additional Qualitative Results}
\label{sec:appendix_qualitative}

\begin{figure}[!ht]
    \centering
    \vspace{-1em}
    \includegraphics[width=\linewidth,height=0.85\textheight,keepaspectratio]{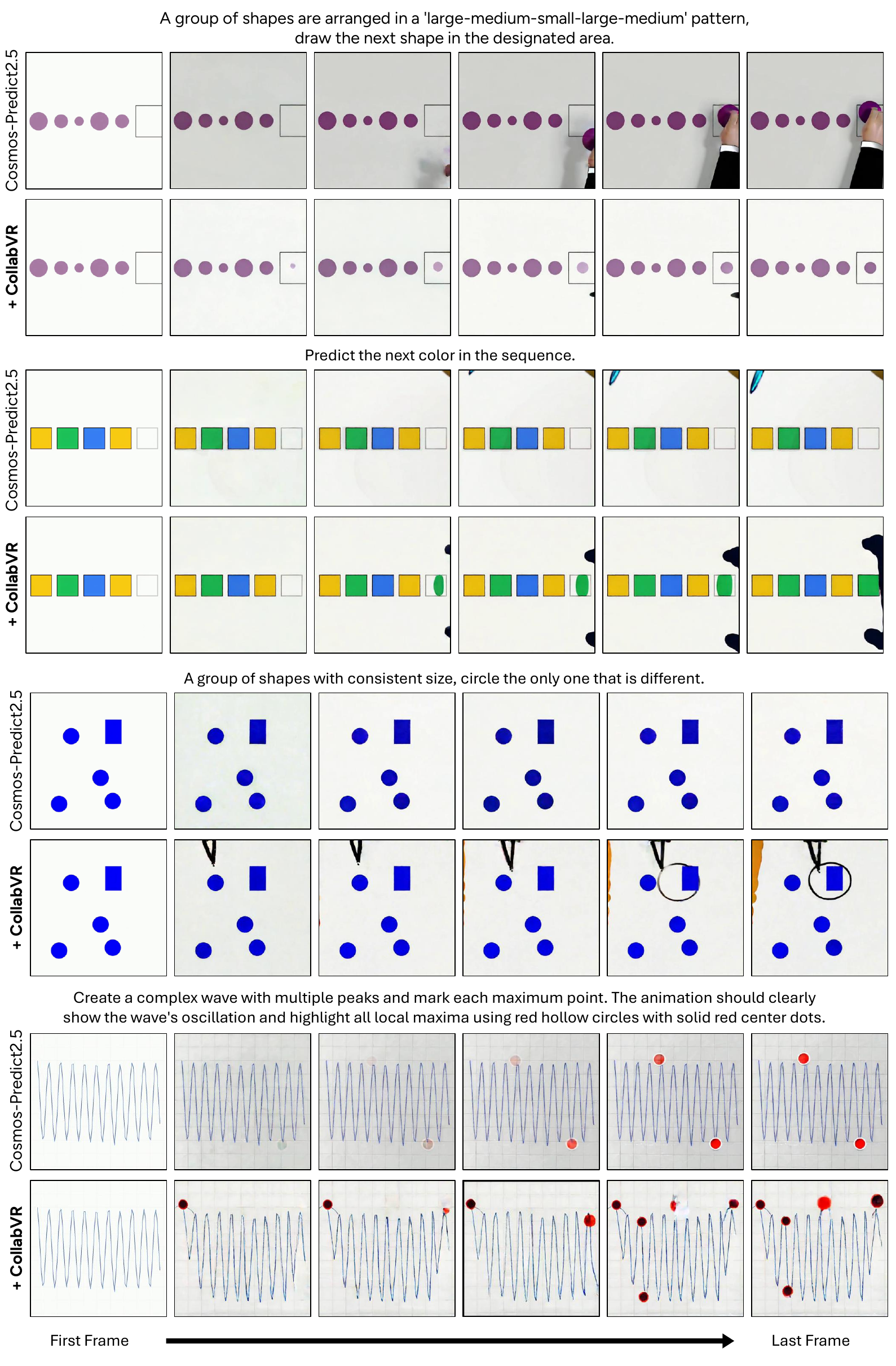}
    \caption{
\textbf{\oursshort{} generalizes to Cosmos-Predict-2.5 on \vbvr{}.}
Each two-row block contrasts Cosmos-Predict-2.5 alone (top) with Cosmos-Predict-2.5+\oursshort{} (bottom) over a six-frame sequence (first $\rightarrow$ last frame).
}
    \label{fig:app:VBVR_Cosmos_by_N}
\end{figure}

\clearpage

\begin{figure}[t]
    \centering
    \vspace{-1em}
    \includegraphics[
    height=0.9\textheight,
    width=\linewidth,
    keepaspectratio
]{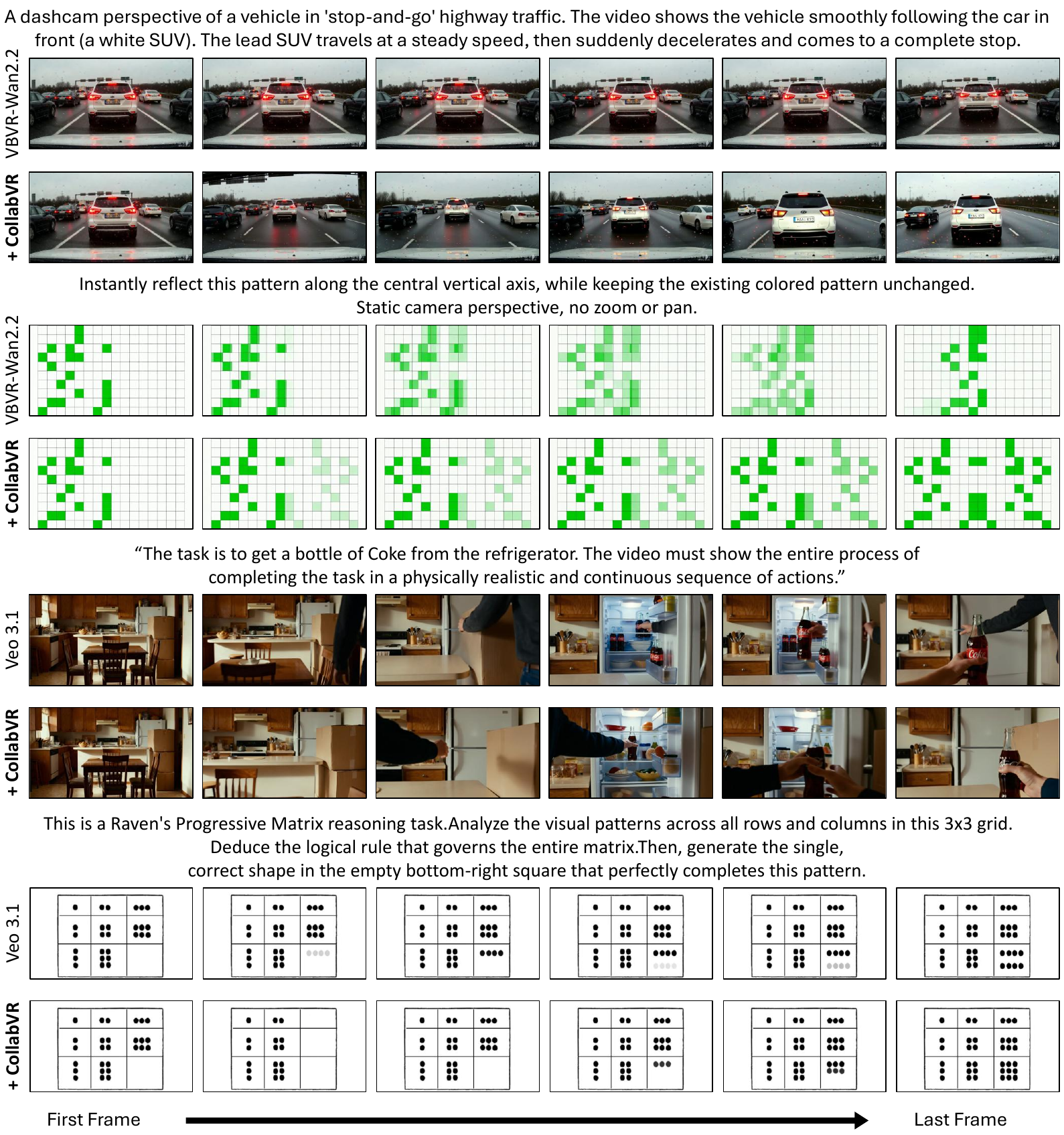}
    \caption{
\textbf{\oursshort{} works across open- and closed-source VGMs on \genvire{}.}
Each two-row block contrasts the base VGM (top) with +\oursshort{} (bottom); the upper two tasks use \vbvrwan{} and the lower two use \veo{}.
}
\vspace{-1em}
    \label{fig:app:VBVR_VGM_GenViRe}
\end{figure}

\subsection{Examples by VGM}
\label{sec:appendix_qual_by_vgm}

\oursshort{} applies to a range of generators beyond \vbvrwan{} through the same interface.

\paragraph{Generalization to another open-source VGM.}
Figure~\ref{fig:app:VBVR_Cosmos_by_N} shows that the same gains carry over to Cosmos-Predict-2.5: although the alone model exhibits its own characteristic failures (e.g., hallucinating an external hand or pen instead of acting on the canvas), pairing it with \oursshort{} consistently drives the generator toward the intended in-canvas action across all four tasks.

\paragraph{Open- and closed-source on \genvire{}.}
Figure~\ref{fig:app:VBVR_VGM_GenViRe} broadens the comparison to \genvire{} with an open-source (\vbvrwan{}) and a closed-source (\veo{}) generator, spanning real-world (dashcam, refrigerator) and symbolic-pattern tasks (mirror reflection, Raven's matrices).
+\oursshort{} produces visibly more faithful executions than the base VGM in every case, supporting the results in Tables~\ref{tab:genvire_results} and~\ref{tab:vbvr_results}.

\clearpage

\begin{figure}[t]
    \centering
    \vspace{-1em}
    \includegraphics[
    height=0.95\textheight,
    width=\linewidth,
    keepaspectratio
]{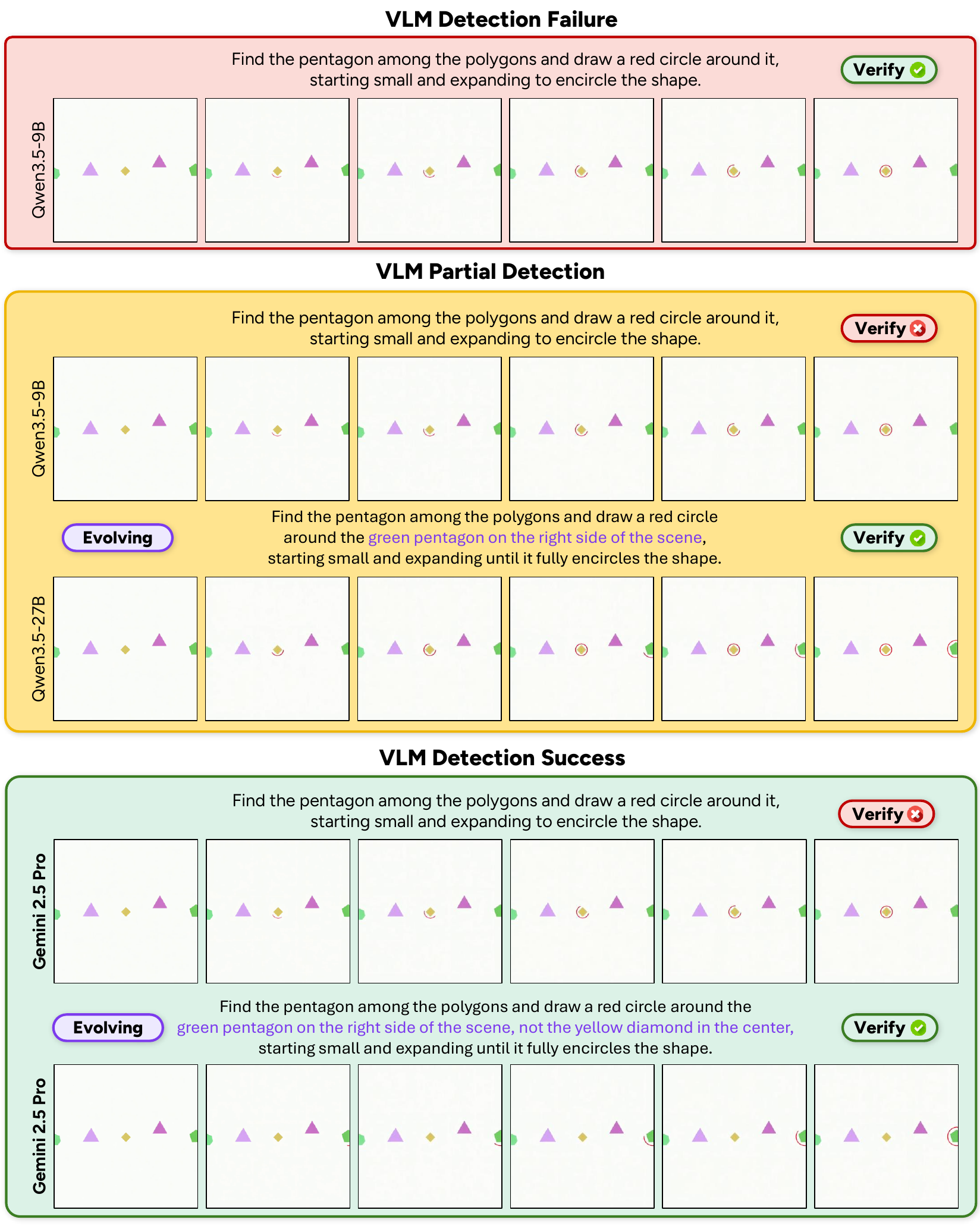}
    \caption{
\textbf{Verifier VLM choice shapes the recovery loop on a single \vbvr{} trace (\vbvrwan{} as the VGM).}
The same prompt is verified by Qwen3.5-9B (top, false-accept), Qwen3.5-27B (middle, evolution with a coarse positional cue), and Gemini~2.5~Pro (bottom, evolution that explicitly excludes the distractor).
}
    \label{fig:app:VBVR_by_VLM_trace}
\end{figure}

\clearpage

\begin{figure}[t]
    \centering
    \vspace{-1em}
    \includegraphics[
    height=\textheight,
    width=\linewidth,
    keepaspectratio
]{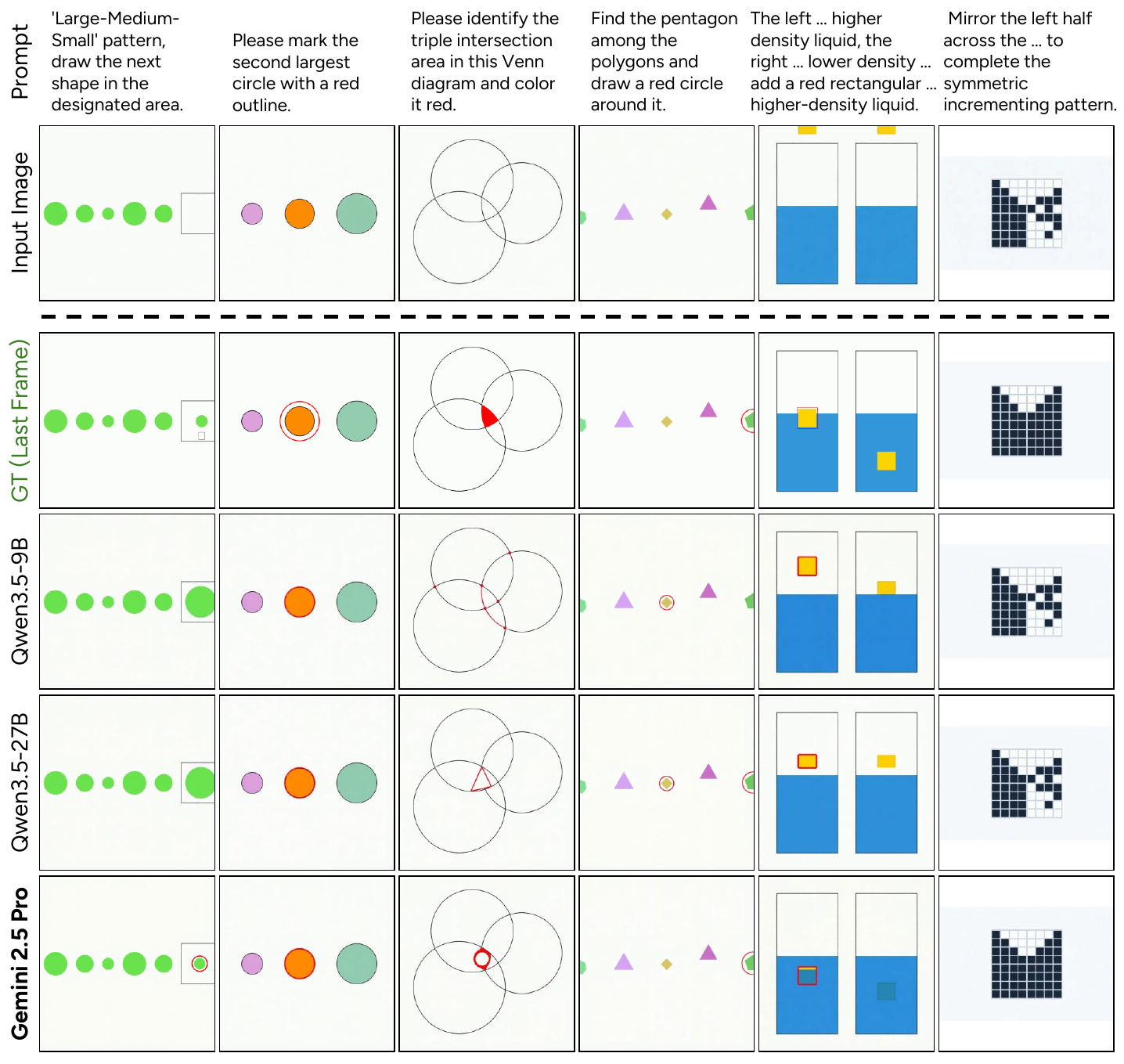}
    \caption{
\textbf{Final +\oursshort{} outputs track verifier capability across \vbvr{} tasks (\vbvrwan{} as the VGM).}
Each row shows the last frame produced when the verifier is Qwen3.5-9B, Qwen3.5-27B, or Gemini~2.5~Pro, with the input image and GT last frame at the top.
}
    \label{fig:app:VBVR_by_VLM_endpoint}
\end{figure}

\subsection{Examples by VLM}
\label{sec:appendix_qual_by_vlm}

The verifier-quality gaps quantified in Section~\ref{sec:human_benchmark} translate into qualitatively distinct recovery behaviors that propagate to the final output.

\paragraph{A single trace under three verifiers.}
Figure~\ref{fig:app:VBVR_by_VLM_trace} traces one regeneration loop on the prompt \textit{``find the pentagon among the polygons and draw a red circle around it''}.
Qwen3.5-9B false-accepts a mis-localized target, so no recovery is triggered.
Qwen3.5-27B rejects but evolves the prompt with only a coarse positional cue (\textit{``on the right side''}), and the regenerated clip is only partially correct.
Gemini~2.5~Pro both rejects and articulates the precise distractor (\textit{``not the yellow diamond in the center''}), and the regenerated clip recovers the correct red circle around the pentagon.

\paragraph{Aggregate behavior across tasks.}
Figure~\ref{fig:app:VBVR_by_VLM_endpoint} confirms this is not a single-trace artifact: across multiple \vbvr{} tasks, proximity of the final +\oursshort{} output to the GT last frame increases monotonically with verifier capability, mirroring the score ordering in Table~\ref{tab:tts_vlm_comparison}.

\begin{figure}[t]
    \centering
    \vspace{-1em}
    \includegraphics[
    height=0.965\textheight,
    width=\linewidth,
    keepaspectratio
]{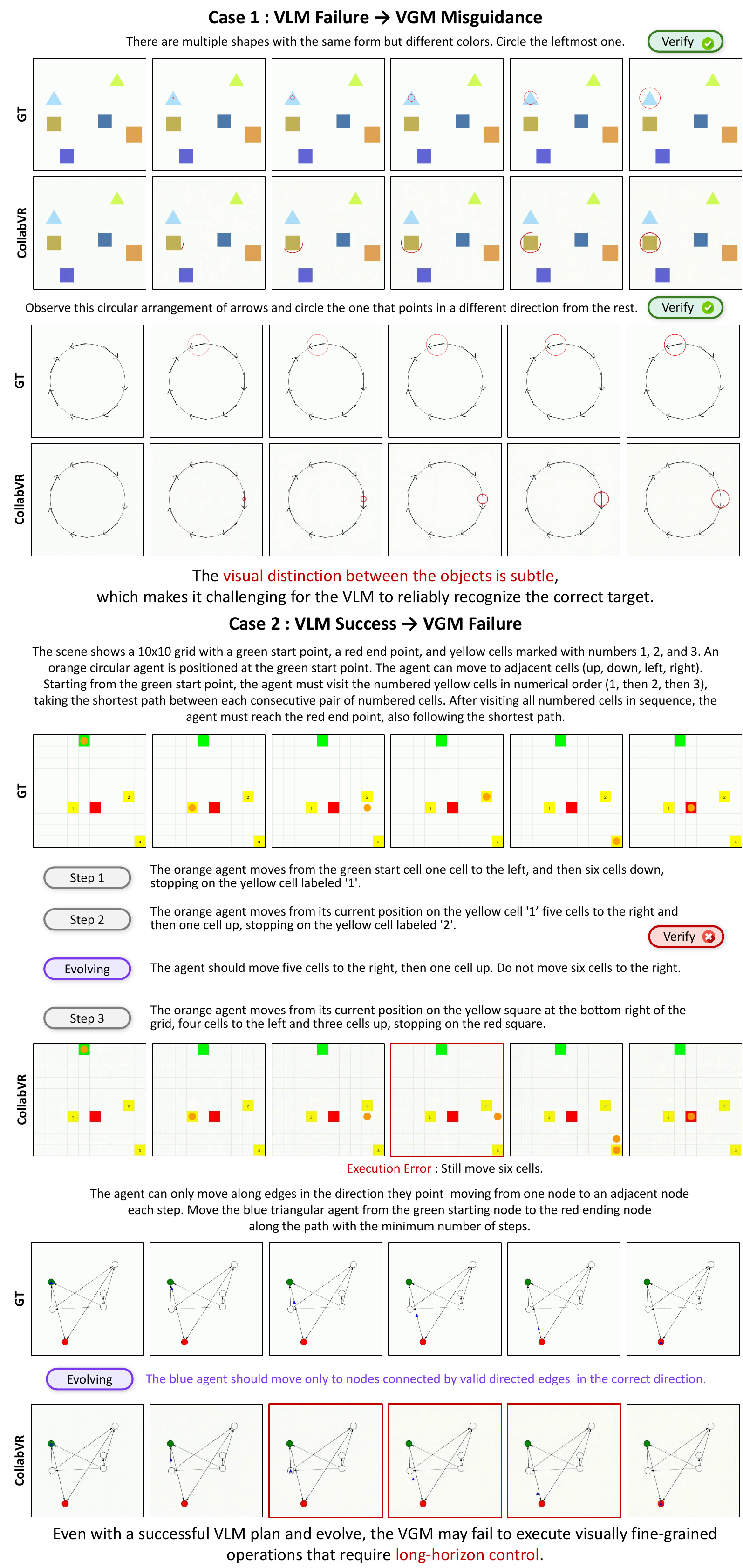}
    \caption{
\textbf{Two distinct ceilings limit full \oursshort{} on \vbvr{}.}
\emph{Case~1}: the VLM verifier fails to detect the issue, so no recovery is triggered.
\emph{Case~2}: the VLM diagnoses the failure and evolves the prompt correctly, but the VGM cannot execute the fine-grained operation.
}
    \label{fig:app:VBVR_Failure}
\end{figure}

\clearpage

\subsection{Failure Cases}
\label{sec:appendix_failure_cases}

Figure~\ref{fig:app:VBVR_Failure} characterizes the residual failures of the full \oursshort{} pipeline, decomposing them into two complementary causes that test-time orchestration alone cannot resolve.

\paragraph{Case~1: VLM detection failure.}
The verifier itself fails to identify the issue, so no recovery is triggered and the incorrect clip is finalized.
\oursshort{} circles a yellow square instead of the leftmost light-blue triangle, and circles an in-distribution arrow instead of the only outward-pointing one---both failures pass the verifier silently.

\paragraph{Case~2: VGM execution failure.}
The verifier correctly diagnoses the failure and the prompt evolves with the right semantic content (\textit{``Do not move six cells to the right.''}, \textit{``valid directed edges in the correct direction''}), yet the VGM still misexecutes the fine-grained operation.
Here the bottleneck has moved from the supervisor to the generator, and additional retries cannot help.

These two ceilings map onto the small-$\Delta$ categories of Figure~\ref{fig:category_module}: the symbolic-category gap (Knowledge, Analogy) is largely a Case~1 issue, while the residual Spatial / Transformation gap reflects a Case~2 ceiling.
They point respectively to stronger VLM grounding and reasoning-oriented VGM training as complementary directions.

\subsection{Partial vs.\ Full Re-generation: Maze Case Study}
\label{sec:maze}

\begin{wrapfigure}{r}{0.52\linewidth}
    \vspace{-1em}
    \centering
    \includegraphics[width=\linewidth]{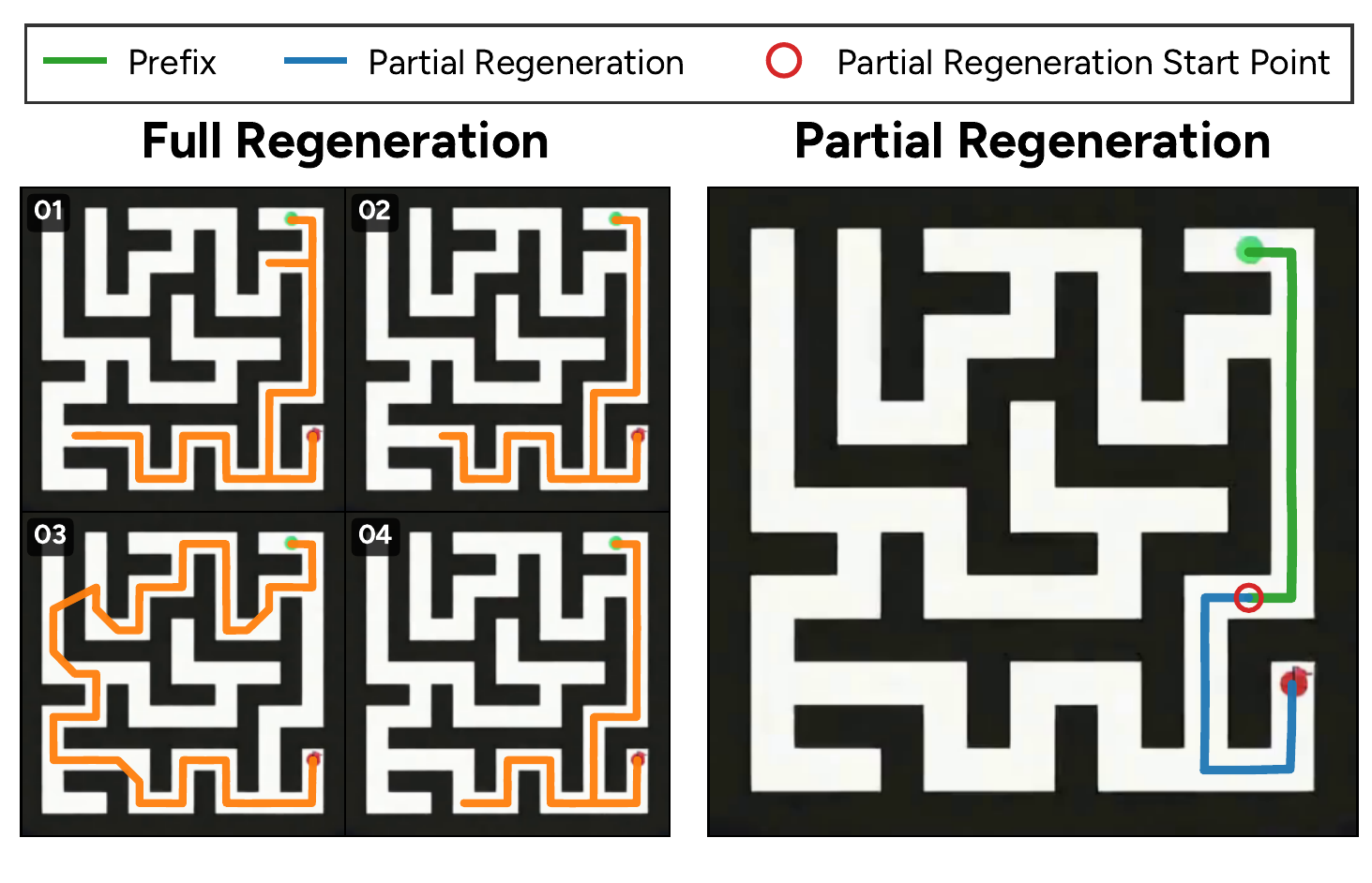}
    \caption{
    \textbf{Partial re-generation from $f_\tau$ outperforms full re-generation on a maze task.}
    Full re-generation is run for four independent attempts (left); partial re-generation reuses the correct prefix up to the first failing frame (right).
    }
    \label{fig:maze_partial_full}
\end{wrapfigure}

We additionally explore a recovery design specific to navigation-style tasks: on rejection, the VGM is re-invoked from the first failing frame $f_\tau$ rather than re-rolling from scratch, so that the prefix already produced correctly is preserved across attempts.

Figure~\ref{fig:maze_partial_full} compares the two modes on a maze instance.
Full re-generation fails to reach the goal across four independent attempts because each resampled trajectory discards previously accumulated progress.
Partial re-generation from $f_\tau$ reuses the correct prefix and successfully converges on the goal cell, illustrating that test-time compute is much more effective when it targets the failed suffix rather than the entire trajectory.

We position partial re-generation as an auxiliary mode applicable when the prefix carries useful information toward the goal, with precise failure detection as a promising future direction.

\section{Broader Impact}
\label{sec:appendix_broader_impact}

\textbf{Positive impacts.}
Video reasoning is becoming a central AI capability for dynamic, temporally-grounded processes that static-image or pure-text reasoning cannot express, with applications spanning educational and scientific demonstrations, procedural walkthroughs, navigation in synthetic environments, and embodied-agent simulation.
\oursshort{} contributes to this direction by showing that strong general-purpose VLMs and VGMs, while individually imperfect for goal-directed video generation, can be composed at step-level granularity into a closed-loop reasoning system that improves task fidelity without any additional training, and that the resulting trajectories are more interpretable and auditable than single-shot VGM outputs because each step's verification decision and prompt evolution are exposed as discrete artifacts.
We see the next frontier of video reasoning coming less from a single ever-larger end-to-end generator than from richer inference-time collaborations between strong specialised models such as VLMs as step-level supervisors, VGMs as visual simulators, and other modality-specific reasoners, with \oursshort{} as one concrete instance of this design pattern.

\textbf{Potential negative impacts.}
The framework inherits the misuse risks of the underlying generative models, including higher-fidelity synthetic video that could be used for deceptive or non-consensual content.
\oursshort{} adds no new generative capability beyond the base VGM; it composes existing models for closed-loop reasoning rather than training new generators.

\textbf{Mitigations.}
We use only off-the-shelf VGMs (Wan2.2, Cosmos-Predict-2.5, Veo 3.1) and VLMs (Gemini 2.5 Pro, Qwen3.5) under their original licenses and existing safeguards.
The released artifacts (orchestration code and the human-annotated reliability benchmark) carry no novel high-risk generative capability beyond what is already publicly available.

\end{document}